\documentclass[10pt,twocolumn,letterpaper]{article}

\usepackage[pagenumbers]{cvpr} %
\usepackage[accsupp]{axessibility}

\usepackage{graphicx}
\usepackage{amsmath}
\usepackage{amssymb}
\usepackage{booktabs}
\usepackage{balance}
\usepackage[T1]{fontenc}

\usepackage{url}
\usepackage{bbm}
\usepackage{tabularx}
\usepackage{algorithm}
\usepackage[noend]{algpseudocode} 
\usepackage{enumitem}
\usepackage{caption}
\usepackage{xspace}
\usepackage{xcolor}
\usepackage{subcaption}
\usepackage{wrapfig,lipsum,booktabs}
\usepackage{amssymb}
\usepackage{comment}
\usepackage{xcolor}

\newtheorem{definition}{Definition}

\newcommand{\our}{PuriDivER\xspace}
\newcommand{\ourfull}{Purity and Diversity aware Episode Replay\xspace}
\newcommand{\ourfulluline}{\underline{Puri}ty and \underline{Div}ersity aware \underline{E}pisode \underline{R}eplay\xspace}

\usepackage[pagebackref,breaklinks,colorlinks]{hyperref}

\usepackage[capitalize]{cleveref}
\crefname{section}{Sec.}{Secs.}
\Crefname{section}{Section}{Sections}
\Crefname{table}{Table}{Tables}
\crefname{table}{Tab.}{Tabs.}

\def\cvprPaperID{10123} %
\def\confName{CVPR}
\def\confYear{2022}

\begin{document}

\title{Online Continual Learning on a Contaminated Data Stream\\ with Blurry Task Boundaries}

\author{
\vspace{0.5em} %
Jihwan Bang$^{1,2}$\hspace{0.5em}Hyunseo Koh$^{2,3}$\hspace{0.5em}Seulki Park$^{2,4}$\hspace{0.5em}Hwanjun Song$^{1,2}$\hspace{0.5em}Jung-Woo Ha$^{1,2}$ \hspace{0.5em}Jonghyun Choi$^{2,5,}$\thanks{: corresponding author.}\\
{$^1$NAVER CLOVA\hspace{1em}
$^2$NAVER AI Lab\hspace{1em}
$^3$GIST\hspace{1em}
$^4$Seoul National University\hspace{1em}
$^5$Yonsei University}\\
{\tt\small {\{jihwan.bang,hwanjun.song,jungwoo.ha\}@navercorp.com, hyunseo8157@gm.gist.ac.kr,}} \\
{\tt\small {seulki.park@snu.ac.kr, jc@yonsei.ac.kr}}
}

\maketitle

\begin{abstract}
Learning under a continuously changing data distribution with incorrect labels is a desirable real-world problem yet challenging. 
A large body of continual learning (CL) methods, however, assumes data streams with clean labels, and online learning scenarios under noisy data streams are yet underexplored. 
We consider a more practical CL task setup of an online learning from blurry data stream with corrupted labels, where existing CL methods struggle.
To address the task, %
we first argue the importance of both diversity and purity of examples in the episodic memory of continual learning models. 
To balance diversity and purity in the episodic memory, we propose a novel strategy to manage and use the memory by a unified approach of label noise aware diverse sampling and robust learning with semi-supervised learning.
Our empirical validations on four real-world or synthetic noise datasets (CIFAR10 and 100, mini-WebVision, and Food-101N) exhibit that our method significantly outperforms prior arts in this realistic and challenging continual learning scenario. Code and data splits are available in \small\url{https://github.com/clovaai/puridiver}.
\end{abstract}

\vspace{-1em}
\section{Introduction}
Continual learning (CL) is a practical learning scenario under a continuous and online stream of annotated data~\cite{lopez2017gem, prabhu2020gdumb, bang2021rainbow}.
Due to continuously changing data distribution, the CL methods are known to suffer from the stability-plasticity dilemma due to catastrophic interference and resistance to learning new information~\cite{mccloskey1989catastrophic,goodfellow2013empirical}.
In addition, when the CL model is deployed in real-world such as e-commerce applications, the annotated labels are often unreliable due to less controlled data curation process, \eg, crowd-sourcing~\cite{xu2021improving}.
Although data labels are likely to be contaminated in real-world due to human errors, existing CL studies have largely assumed that the given training data has no annotation error~\cite{kirkpatrick2017ewc, prabhu2020gdumb}, which might hinder practical usages of many CL methods in real-world applications.

A very recent work~\cite{kim2021continual_spr} relaxes this assumption by proposing an online CL setup under data stream with less reliable labels.
However, they assume a disjoint class incremental scenario such that there is no class overlapping between task streams, which is argued as {less} practical for real-world applications in the literature~\cite{prabhu2020gdumb,bang2021rainbow}.
To step forward in addressing continuously changing data distribution that can be falsely labeled, we first design a novel CL task of \textit{online continual learning on a contaminated data stream on blurry task boundaries~\cite{prabhu2020gdumb}} as a more realistic and practical continual learning scenario. 

Sample selection strategy is arguably important for episodic memory-based CL methods \cite{kirkpatrick2017ewc, prabhu2020gdumb, bang2021rainbow, aljundi2019online, kim2021continual_spr}. 
Specifically, diversity-aware sample selection policy is shown to be effective in the online blurry CL scenarios~\cite{bang2021rainbow}. 
However, on the contaminated data stream, diversity-based memory construction may promote to include many examples with corrupted labels in the episodic memory, leading to poor performance thanks to very high capacity of deep models to overly fit all the falsely labeled examples~\cite{zhang2021understanding}.

\begin{figure*}[t!]
\centering
\includegraphics[width=0.75\textwidth]{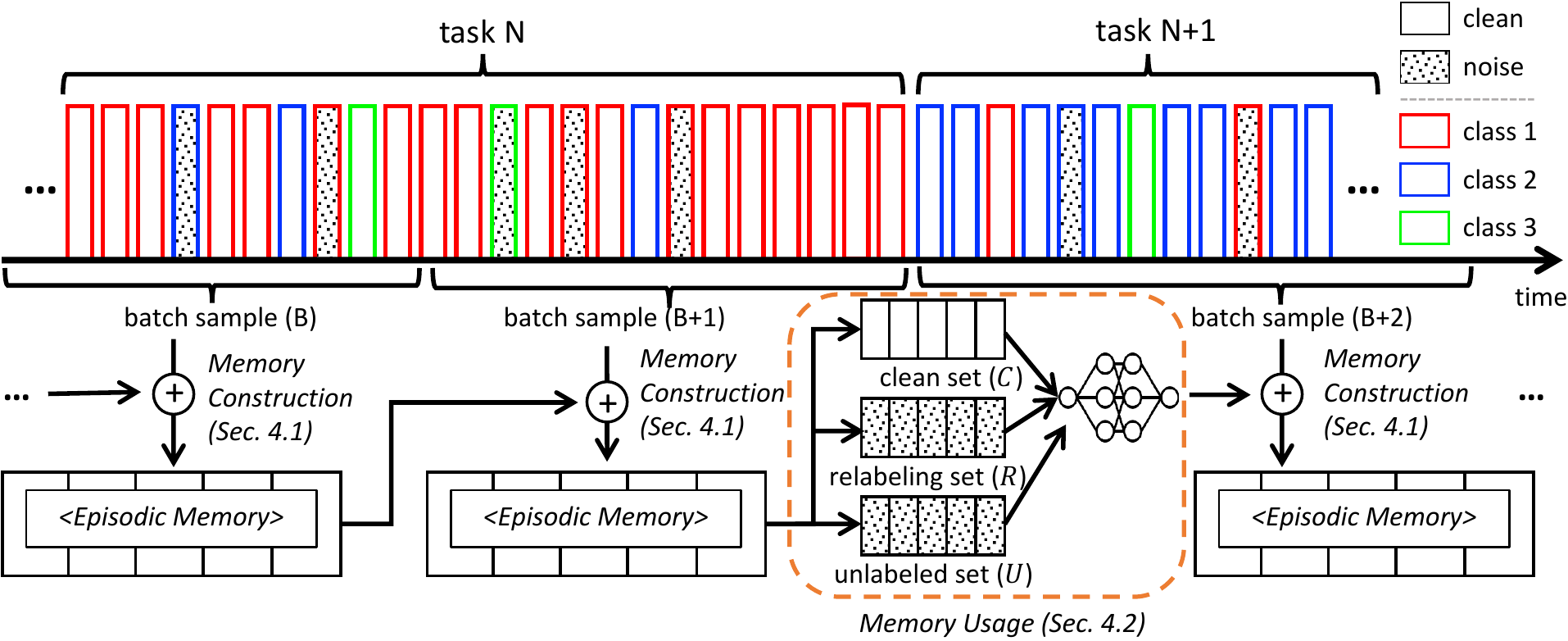}
\vspace*{-0.1cm}
\caption{Overview of \our for the online blurry continual learning with noisy labels, where all the tasks share classes (\ie, blurry tasks~\cite{prabhu2020gdumb}) and examples might be falsely labeled. \our updates episodic memory by balancing examples' diversity and purity, and unifies a robust training scheme based on semi-supervised learning not to be interfered by unreliable labels in the episodic memory.
}
\label{fig:overview}
\vspace{-1em}
\end{figure*}

To address this issue, we propose a unified framework of memory sampling policy with robust learning with the episodic memory and semi-supervised learning with unreliable data, named \textbf{\our} (\ourfulluline). 
Fig.~\ref{fig:overview} illustrates an overview of the proposed method to address the new CL task set-up.

Specifically, for the memory sampling policy, we propose to construct a robust episodic memory that preserves a set of training examples that are \emph{diverse} and \emph{pure}.
Unfortunately, maintaining the information diversity and purity is in \emph{trade-off}; clean examples mostly exhibit smaller losses (less diverse) than noise examples due to the memorization effect of deep neural networks~\cite{arpit2017closer, song2021robust}.
Thus, emphasizing purity in memory sampling does not promote sample diversity and \emph{vice versa}. %
To address the dilemma, we define a score function of promoting the label purity with an additional term to promote diversity by optimizing the sample distribution to be similar to the one with noisy examples. %
In addition, we incorporate a novel robust learning scheme to further promote both purity and diversity when we use the samples from the episodic memory (Sec.~\ref{sec:sub:robustlearning}).
Thus, we enforce high diversity and purity around the memory twice; by its construction and usage.

Our empirical validations show that the proposed method outperforms combinations of prior arts to address the combined challenges by noticeable margins on multiple evaluations using CIFAR-10/100, WebVision and Food-101N.
Our contributions are summarized as follows:
\vspace{-0.5em}
\begin{itemize}[leftmargin=10pt]
\setlength\itemsep{-0.5em}
    \item Proposing the first online \emph{blurry} CL set-up with contaminated data stream, which is realistic and challenging, and establishing a strong baseline for it.
    \item Proposing a unified framework of considering diversity and purity in memory updates and usage strategy.
    \item Proposing an adaptive balancing scheme for dynamic trade-off between diversity and purity in memory sampling for addressing various noise ratio in CL.
    \item Proposing a semi-supervised robust learning scheme with label cleaned up samples and unreliable samples regarded as unlabeled samples.
\end{itemize}

\section{Related Work}
\label{sec:related_work}

There is a large body of literature to address the two challenges of online CL and learning from noisy data, separately. 
Here, we selectively review the most relevant ones.

\subsection{Task-free Blurry Online Continual Learning}

Recent studies~\cite{bang2021rainbow,prabhu2020gdumb,aljundi2019gss} argue that the disjoint CL setup is less realistic but prevalent in literature and propose a new setup where task boundaries share classes, called `blurry'.
While the disjoint setup assumes each task has non-overlapping classes, the blurry setup allows that all tasks possibly share the same set of classes with different class distributions. 
Moreover, we categorize CL methods by the way of using the data; online and offline.
Offline setup allows a model to use data many times, while online setup only allows a model to observe a data stream once at the current moment, which is more difficult but practical. %

Task-free setup refers to the setup that does not allow the access of task id at inference.
There have been many studies on task-free online continual learning (either blurry or disjoint), which are categorized into two; \textit{1) rehearsal-based approaches}~\cite{bang2021rainbow,prabhu2020gdumb,riemer2018learning}, where episodic memory is used to store a few examples of old tasks, and \textit{2) regularization-based approaches}~\cite{kirkpatrick2017ewc,rwalk}, which regularizes neural network parameters from drastic updates for new information to preserve the information of old tasks for less forgetting.
Since rehearsal-based approaches generally outperform the regularization ones, we use it. %

For better rehearsal with an episodic memory, many strategies have been proposed for data sampling.
Unfortunately, most of them are not suitable for the online continual learning setup except the following few.
MER~\cite{riemer2018learning} applied \textit{reservoir sampling} for updating memory in online continual learning, and it can approximately optimize over the stationary distribution of all seen examples. GDumb~\cite{prabhu2020gdumb} proposed the \textit{greedy balancing sampler} such that it focuses on balancing the number of examples for each class. 
\textit{Rainbow memory}~\cite{bang2021rainbow} and OCS~\cite{yoon2021online} are proposed to diversified sampling by regular interval of uncertainty and constructing a `Core-Set'~\cite{sener2018coreset}, respectively.
However, proposed methods assume reliable labels only.

\subsection{Learning with Unreliable Labels}
A number of approaches have been proposed to prevent a model from overfitting to unreliable or noisy labels. 
One of the typical methods is `sample selection', which trains the model for selected small-loss examples whose labels are likely to be correct. MentorNet~\cite{jiang2018mentornet} maintains a teacher network to provide the student network with the small-loss examples. Similarly, Co-teaching(+) \cite{han2018coteaching, yu2019coteachingplus} utilizes two models, where each model selects a certain number of small-loss examples and feeds them to its peer model for training. 
Alternatively, `re-labeling' corrects noisy labels to reliable ones. SELFIE \cite{song2019selfie} re-labels the examples by predicted labels' consistency. SEAL \cite{chen2021seal} averages the model's softmax output of each example over the whole training set. 

Although they utilize the entire data, when the number of classes or noise rate increases, a risk of overfitting to false labels still persists due to possible false correction. %
To address this, there have been efforts to combine `semi-supervised learning' with `sample selection.' 
SELF \cite{nguyen2019self} progressively filters out falsely labeled examples from noisy data by using self-ensemble predictions. DivideMix \cite{berthelot2019mixmatch} fits bi-modal univariate Gaussian mixture models\,(GMMs) to the loss of training examples for sample selection, and then applied the MixMatch algorithm \cite{li2019dividemix}. 
However, this family of robust methods assumes the offline setup with sufficient amount of training data whereas the training data is limited in the blurry online continual learning setup.

\subsection{Continual Learning with Unreliable Labels}

A straightforward approach for CL with unreliable labels is combining a memory sampling method for online learning with a robust learning method for handling the unreliable labels. 
However, this naive integration may not resolve the combined challenge because of aforementioned trade-off between diversity and purity of the examples.
Recently, SPR~\cite{kim2021continual_spr} is proposed to solve online continual learning with contaminated data via self-purifying scheme. 
However, the experiments are only conducted on disjoint setup that is less realistic, and there are large rooms to improve in accuracy even in that setup (see Sec. 5 in the supplementary material).
In contrast, we approach this problem by constructing the episodic memory with high diversity and purity, and mitigate the potential risk of memory sampling by using a complementary robust learning approach (see Tab.~\ref{tab:dpm_ablation}) in more realistic CL scenario.

\section{Task Definition}
\label{sub:problem}

We design a new online CL setup that is on a contaminated data stream with \emph{blurry} task boundaries following \cite{bang2021rainbow}.
Let $C$ and $T$ be the number of classes and tasks, respectively. 
At each task $t$, the set of classes are split into a set of \emph{major classes}, $M_t$ and a set of \emph{minor classes}, $m_t$. 
For describing the imbalanced degree of major and minor classes, we define the hyperparameter ($L$) as the number of minor classes over the number of all the classes, and those tasks are called \textit{blurry-$L$}.
Each class could be either a major class once in a certain task or a minor class in the other tasks. 
Formally, tasks should satisfy the following conditions:
\vspace{-0.5em}
\begin{equation}
\begin{split}
& \cup_{t=1}^{T} M_t = \cap_{t=1}^{T} M_t \cup m_t = \{1, \dots ,C\}, \\ 
& \cup_{i, j = 1, i \neq j}^{T}M_i \cap M_j = \varnothing.
\vspace{-1em}
\end{split}
\end{equation}

To configure blurry tasks, we split the streaming data $\mathcal{S}_t$ for the task $t$ into subsets of different classes; $\mathcal{S}_t = \{\mathcal{S}_{(t, c)} : c \in \{1, \dots ,C\} \}$. 
For a given task $t$, $L = |\cup_{c \in m_t} \mathcal{S}_{(t, c)}| / |\mathcal{S}_t|$, where $L$ is the specified class imbalance ratio at task $t$. 
In particular, in the presence of noisy labels, the streaming data {\small$\mathcal{S}_t$} for a task $t$ include falsely labeled examples, thus {\small $\mathcal{S}_t=\{(x_i, \tilde{y_i})\}_{i=1}^{|\mathcal{S}_t|}$} where $x_i$ is an example and $\tilde{y_i}$ is its noisy label which may be different from the true (\ie, not contaminated) label $y_i$.

\section{Pure and Diverse Episode Replay}

To address the online blurry continual learning with unreliable labels, we propose a method to construct and use the episodic memory.
Specifically, we propose (1) a sampling method `diverse' exemplars from the online stream for episodic memory with label purifying scheme (Sec. \ref{sub:memory_update}), and (2) to use the `purified' diverse examples from the memory by semi-supervised learning with noisy labels for extra diversity with purity (Sec. \ref{sec:sub:robustlearning}).
We call this method \ourfull (\textbf{\our}).
We describe a brief version of algorithm in Alg.~\ref{alg:abstract} and detailed version in the supplementary material (Sec. 1).

\setlength{\textfloatsep}{7pt}%
\begin{algorithm}[t]
\small
    \caption{\ourfull}
    \label{alg:abstract}
    \begin{algorithmic}[1]
        \State {\textbf{Input:} $\mathcal{S}_{t}$: stream data at task $t$, $\mathcal{M}$: exemplars stored in a episodic memory, $T$: the number of tasks, $\theta_0$: initial model.}
	    \For {$t$ = 1 {\bf to} T}
	        \For {each mini-batch $\mathcal{B} \in \mathcal{S}_{t}$}

	       \State{$\theta \leftarrow \theta - \alpha \nabla \sum_{i \in \mathcal{B}} \ell(\theta(x_i), \tilde{y_i})$} {\small\color{orange}\Comment{online training}}
	            \For {$x_i, \tilde{y_i}$ in $\mathcal{B}$}
	            \State{Update $\mathcal{M}$ from $\mathcal{M} \cup (x_i, \tilde{y_i})$}{\small\color{orange}\Comment{Sec.~\ref{sub:memory_update}}}
	        \EndFor
	   \EndFor
	   \For {$e=1$ {\bf to} MaxEpoch}
	        \State{Split $\mathcal{C}$, $\mathcal{R}$, $\mathcal{U}$ from $\mathcal{M}$ via  Eqs.~\eqref{eq:gmm_loss} and \eqref{eq:uncertainty_loss}}
	        \For {each mini-batch $\mathcal{B_C} \in \mathcal{C}$, $\mathcal{B_R} \in \mathcal{R}$, $\mathcal{B_U} \in \mathcal{U}$ }
	            \State{$\theta \leftarrow \theta - \alpha \nabla [\ell_{cls} + \eta \ell_{reg}$]}{\small\color{orange}\Comment{Sec.~\ref{sec:sub:robustlearning}}}
	        \EndFor
	   \EndFor
	   \EndFor
    \end{algorithmic}
\end{algorithm}

\subsection{Episodic Memory Construction}
\label{sub:memory_update}
Selecting a diverse set of examples for the episodic memory is arguably one of the beneficial steps to improve accuracy of CL model~\cite{bang2021rainbow}. 
Since labels are unreliable in our setup, however, this approach would expedite overfitting to falsely labeled examples because they are likely regarded as the ones that are highly diverse thus included in the memory and used in training. 
On the other hand, to address purity, we may use a small-loss trick for purity aware sampling, but it only selects the examples with low diversity.
Thus, the diversity and purity are in trade-off.

To address the dilemma, we define a sampling score function that considers both aspects for the episodic memory, and it is lower the better.
Specifically, the proposed score function consists of two factors; (1) purity of an example by assuming that the `small-loss examples' are likely correctly labeled, and (2) the similarity in representation space between the `relevant representation' of the sample to all other samples with the same noisy label in the memory for diversity. %
The representation vector is defined as:
\begin{definition}
\label{definition_relevant}
Let {$f(x_i) \in \mathbbm{R}^n$} be the representation vector of an example $x_i$, where $f(\cdot)$ be a representation learner. A \textbf{relevant representation} {$f_{rel}(x_i; j)$} for a class $j$ is defined as a subset of elements {$e (\in f(x_i)$)} such that {\small $f_{rel}(x_i; j)=\{ e \in f(x_i) ~|~ w{(e,j)} > \frac{1}{|C|} \sum_{c=1}^{C}w{(e,c)}\}$}, where {$w{(e,c)}$} is a weight parameter of the classification FC layer associated with the element $e$, its inputs to the FC layer, and class $j$.
\label{def:relevant_feature}
\end{definition}

Formally, we define and use the score function of an example $x_i$ to select samples, considering both the purity and diversity with its noisy label $\tilde{y_i}$ as:
\begin{equation}
\small
\label{eq:score}
\begin{aligned}
S(&x_i, \tilde{y_i})  = (1-\alpha_k) \overbrace{\ell\big(x_i, \tilde{y_i} \big)}^{\text{purity}} \\
 & + \alpha_k \overbrace{ \frac{1}{\big|\mathcal{M}[\tilde{y_i}]\big|} \sum_{ \hat{x_i} \in \mathcal{M}[\tilde{y_i}]}{\rm cos}\big(f_{rel}(x_i; \tilde{y_i}), f_{rel}(\hat{x_i}; \tilde{y_i})\big)}^{\text{diversity}},
\end{aligned}
\end{equation}
where $\mathcal{M}$ is the episodic memory at the current moment, $k$ is minibatch index to which $x_i$ is belong and $\mathcal{M}[\tilde{y_i}]$ is a set of examples annotated as $\tilde{y_i}$ in the memory. 
We use the cosine similarity (\ie, $\cos(\cdot, \cdot)$) to measure the similarity between two representations. 
$\alpha_k$ is the balancing coefficient of diversity and purity.
When the score is high, the sample is more likely corrupted in label and does not contribute to diversity much.

Specifically, the high value of the first term (for purity) implies that $\tilde{y_i}$ can be a corrupted label with high probability. 
The high value of the second term (for diversity) implies that the examples annotated with $\tilde{y_i}$ in the memory are similar to $x_i$, which can have less impact to train the model.
By the computed score, we update the memory by dropping a single example with the highest score when we add a new sample from the data stream.
Finding a proper $\alpha_k$ is another non trivial issue. %

\vspace{-1em}
\paragraph{Adaptive Balancing Coefficient.}
We argue that the optimal balancing coefficient $\alpha_k$ could vary depending on the label noisiness or learning difficulty of each example\,(Sec.~\ref{sec:coefficent} for empirical analysis).
To make the algorithm adaptive to learning difficulty and noisiness of an example, we additionally propose an adaptive $\alpha_k$. %

As argued in curriculum learning literature~\cite{bengio2009curriculum,graves2017automated}, easy or clean examples are favored in the early stage of training of a model. 
Similarly, we conjecture that diverse examples are more important at the later stage of training for better generalization.
To implement this idea that implicitly considers learning difficulty and noise ratio, we propose an adaptive balancing coefficient for a $k^{th}$ minibatch based on the example's training loss that is a proxy of label noisiness and learning difficulty (\ie, if loss is large, learning that sample is difficult) as:
\begin{equation}
    \alpha_k = 0.5 \cdot \min \left( 1 / {\ell(\mathbf{X}_k,\mathbf{\tilde{Y}}_k)}, 1 \right),
\end{equation}
where $\ell(\mathbf{X}_k,\mathbf{\tilde{Y}}_k)$ is the average loss for all examples in the $k^{th}$ batch. 
We empirically validate its effectiveness in Sec.~\ref{sec:coefficent}.

\subsection{Episodic Memory Usage}
\label{sec:sub:robustlearning}

Even with the memory of diverse and expectedly pure samples, the episodic memory may include noisy examples, since the label corrupted examples are mostly one with high diversity. %
However, due to the small size of the episodic memory, robust learning with the memory does not bring significant accuracy improvement.
Here, we propose a hybrid learning scheme that combines re-labeling and unsupervised learning with unreliably labeled examples.

Specifically, we first split the examples in the episodic memory into clean and noisy subsets based on the small-loss trick that is widely used when handling noisy labels~\cite{li2019dividemix, arazo2019betamixture}. 
We then fit a GMM to the training loss of all examples in the memory such that it estimates the probability of an example being $p_G(\cdot)$~\cite{li2019dividemix}. 
Given a noisy example $x_i$ with its unreliable label $\tilde{y_i}$, we obtain probability of label purity by the posterior probability of GMMs as:
\begin{equation}
p_{G}\big(g | \ell(x_i, \tilde{y_i}; \Theta)\big) = \frac{p_{G}\left(\ell(x_i, \tilde{y_i};\Theta) | g \right) \cdot p_{G}(g)}{p_{G}\left(\ell(x_i, \tilde{y_i}; \Theta) \right)},
\end{equation}
where $g$ denotes a Gaussian modality for the small-loss examples, $\Theta$ denotes weight parameters of the model, $\ell(x_i,\tilde{y_i};\Theta) = - \log(p_{m} (x_i, \tilde{y_i}; \Theta))$ and $p_{m}(x_i, c)$ is the softmax output of the model for class $c$. 
We finally obtain the clean set $\mathcal{C}$ and noisy set $\mathcal{N}$ as:
\begin{equation}
\begin{gathered}
\mathcal{C} := \big\{ (x_i, \tilde{y_i}) \!\in\! \mathcal{M} : p_{G}\big(g | \ell(x_i, \tilde{y_i}; \Theta)\big) \!\geq \!0.5\big\}, \\
\mathcal{N} := \big\{ (x_i, \tilde{y_i})\! \in\! \mathcal{M} :  p_{G}\big(g | \ell(x_i, \tilde{y_i}; \Theta)\big) \!< \!0.5\big\}.
\label{eq:gmm_loss}
\end{gathered}
\end{equation}

In modern approaches of handling noisy labels~ \cite{han2018coteaching, yu2019coteachingplus, song2021robust}, only the clean set is used to train the model in a supervised manner, while the noisy set is discarded for robust learning. 
However, in continual learning, as an episodic memory only contains a few number of samples, we need to utilize as many samples as possible, instead of discarding any.
A straightforward way of using them is to purify them.

However, considering that the memory contains samples with various classification confidences (\ie, logit values of a model), if we use the low confidence examples for re-labeling, they are likely incorrectly re-labeled (\ie, purified) due to the perceptual consistency of a neural net.~\cite{reed2015bootstrap}. 

To address this issue, we further split the noisy set into two subsets; one with high confidence to be re-labeled and the other with low confidence to be used for unsupervised learning. 
Then, we re-label the former but consider the latter as an unlabeled examples and employ the unsupervised learning; specifically, we use `consistency regularization.'

In particular, to split the noise set into the two sets, we define \emph{predictive uncertainty} as:
\begin{equation}
\label{eq:uncertainty}
{\rm U}(x_i) = 1.0 - \max_{c} \big(p_{m}(x_i, c; \Theta)\big),
\end{equation}
where $c \in \{1, \dots ,C\}$.
According to our empirical analysis (see Fig.~\ref{fig:validity_gmm_fitting}-(b)), the uncertainty of the two sets follows \emph{bi-modal} distributions similar to the loss distribution of clean and noise examples. To distinguish two modalities without knowing ground-truth labels, 
we again fit a GMM to the uncertainty of the (expected) noisy examples in the noise set $\mathcal{N}$. 
Similar to Eq.~\ref{eq:gmm_loss}, we split the noise set into the `re-labeling set' ($\mathcal{R}$) and the `unlabeled set' ($\mathcal{U}$) as: 
\begin{align}
&\mathcal{R} := \big\{ (x_i, \tilde{y_i}) \!\in\! \mathcal{N} : p_{G}\big(u | {\rm U}(x_i)\big)\big) \!\geq \!0.5\big\} \nonumber, \\
&\mathcal{U} := \big\{ (x_i, \tilde{y_i})\! \in\! \mathcal{N} :  p_{G}\big(u | {\rm U}(x_i)\big) \!< \!0.5\big\},
\label{eq:uncertainty_loss}
\end{align}
where $u$ is the Guassian component for the low-uncertainty examples. 
Since the examples with high uncertainty are difficult to estimate the true label, we treat them as the unlabeled set instead to prevent false supervision~\cite{rizve2020uncertainty_aware}.

\vspace{-1em}\paragraph{Re-labeling on $\mathcal{R}$.}
We refurbish the example with low uncertainty (\ie, high confidence) by the convex combination of its given noisy label $\tilde{y_i}$ and the model's prediction $p_{m}(x_i)$, whose coefficient is its confidence, \ie, the posterior probability of the mixture component $u$ as:
\begin{equation}
\label{eq:relabel}
\hat{y_i} =  p_{G}(u | {\rm U}(x_i)) \cdot  p_{m}(x_i) + \big(1.0-p_{G}(u | {\rm U}(x_i))\big) \cdot \tilde{y_i}.
\end{equation}

Employing a soft re-labeling approach to progressively refine the given noisy label based on the model's prediction evolving during training, we mitigate the overfitting to incorrectly modified labels at the beginning of training. 
Hence, the training loss for the clean set ($\mathcal{C}$) and re-labeling set ($\mathcal{R}$) is formulated with their label supervision as:
\begin{equation}
\!\!\ell_{cls} = \frac{1}{|\mathcal{C} \cup \mathcal{R}|} \cdot (\!\!\!\!\!\sum_{(x_i, \tilde{y}_i) \in \mathcal{C}} \!\!\!\!\!\ell(x_i, \tilde{y}_i; \Theta) + \!\!\!\!\!\!\sum_{(x_j, \hat{y}_j) \in \mathcal{R}}\!\!\!\!\!\ell(x_j, \hat{y}_j; \Theta)\big).\!
\end{equation}

\vspace{-1em}
\paragraph{Semi-supervised Learning with $\mathcal{U}$.}
We finally learn information from less confident samples in $\mathcal{U}$.
As their labels are very unreliable, relabeling might not correct them. 
Thus, we consider them as unlabeled data and use consistency regularization, which is widely used in the semi-supervised learning literature\,\cite{berthelot2019mixmatch, tarvainen2017mean}. 
It helps learn useful knowledge from the examples without their labels by penalizing the prediction difference between the two examples transformed from the same one as:
\begin{equation}
\label{eq:ssl}
\ell_{reg} = \frac{1}{|\mathcal{U}|} \sum_{x_i \in \mathcal{U}} \big|\big|p_{m}\big(s(x_i);\Theta\big) - p_{m}\big(w(x_i);\Theta\big)\big|\big|_2,
\end{equation}
where $s(\cdot)$ and $w(\cdot)$ are strong and weak augmentation functions for an example $x_i$. 
We use AutoAugment\,\cite{cubuk2019autoaugment} as $s(\cdot)$ and random horizontal flipping as $w(\cdot)$.

Thus, our final loss function on the three meticulously designed subsets, namely a clean set $\mathcal{C}$, a re-labeled set $\mathcal{R}$, an unlabeled set $\mathcal{U}$, and $\mathcal{M}=\mathcal{C} \cup \mathcal{R} \cup \mathcal{U}$, can be written as:
\begin{equation}
\label{eq:final}
\ell(\mathcal{M}) = \ell_{cls}(\mathcal{C} \cup \mathcal{R}) + \eta \cdot \ell_{reg}(\mathcal{U}).
\end{equation}

\section{Experiments}

\subsection{Experimental Set-up}
\label{sec:exp_setup}

\paragraph{Datasets.} 
We empirically validate our \our on image classification tasks on the blurry CL setups~\cite{bang2021rainbow} with corrupted labels.
We use CIFAR-10, CIFAR-100\,\cite{cifar}, a real-world noisy data of large-scale crawled food images; Food-101N~\cite{lee2018cleannet_food101n}, and a subset of real-world noisy data consisting of large-scale web images whose classes are same as ImageNet~\cite{deng2009imagenet}; WebVision~\cite{li2017webvision}.

\vspace{-1em}\paragraph{Noisy Label Set-ups.} 
Following~\cite{li2019dividemix, han2018coteaching, yu2019coteachingplus, song2019selfie}, we inject two types of synthetic noise that is widely used in the literature to emulate noisy label in CIFAR-10/100; namely symmetric\,(SYM) and asymmetric\,(ASYM) label noise. 
Symmetric noise flips the ground-truth (GND) label into other possible labels with equal probability, while asymmetric noise flips the GND into one with high probability. 

For thorough evaluations, we adjust the ratio of label noise from 20\% to 60\%. 
We use the real label noise in Food-101N and WebVision datasets. 
According to \cite{song2020learning}, estimated noise ratio of both datasets are $20.0\%$. 
We provide the detailed experimental configurations for online CL with noisy labels in the supplementary material (Sec. 2).

\begin{table*}[t!]
\centering
\caption{Last test accuracy on CIFAR-10 (K=500) and CIFAR-100 (K=2,000) with SYM-\{20\%, 40\%, 60\%\} and ASYM-\{20\%. 40\%\}, where $K$ denotes size of episodic memory.}
\vspace*{-0.2cm}
\label{tab:cifar_result}
\resizebox{1.0\linewidth}{!}{%
\begin{tabular}{@{}lrrrrrrrrrr@{}}
\toprule 
& \multicolumn{5}{c}{CIFAR-10} & \multicolumn{5}{c}{CIFAR-100} \\
\cmidrule(lr){2-6}  \cmidrule(lr){7-11} 
{Methods}& \multicolumn{3}{c}{Sym.} & \multicolumn{2}{c}{Asym.} & \multicolumn{3}{c}{Sym.} & \multicolumn{2}{c}{Asym.} \\ 
& \multicolumn{1}{c}{20} & \multicolumn{1}{c}{40} & \multicolumn{1}{c}{60} & \multicolumn{1}{c}{20} & \multicolumn{1}{c}{40} & 
\multicolumn{1}{c}{20} & \multicolumn{1}{c}{40} & \multicolumn{1}{c}{60} & \multicolumn{1}{c}{20} & \multicolumn{1}{c}{40} \\

\cmidrule(lr){1-1} \cmidrule(lr){2-4} \cmidrule(lr){5-6} \cmidrule(lr){7-9} \cmidrule(lr){10-11}
RSV~\cite{riemer2018learning}  & 54.5 \footnotesize{$\pm$ 2.1} & 39.2 \footnotesize{$\pm$ 0.9} & 28.7 \footnotesize{$\pm$ 0.4} & 53.6 \footnotesize{$\pm$ 1.6} & 40.0 \footnotesize{$\pm$ 1.2}  
           & 29.4 \footnotesize{$\pm$ 0.1} & 19.3 \footnotesize{$\pm$ 1.2} & 10.5 \footnotesize{$\pm$ 0.4}  & 29.8 \footnotesize{$\pm$ 0.8} & 20.3 \footnotesize{$\pm$ 1.0} \\ 
~~ $+$ SELFIE~\cite{song2019selfie} & 54.5 \footnotesize{$\pm$ 1.7} & 39.2 \footnotesize{$\pm$ 1.1}& 28.8 \footnotesize{$\pm$ 2.9} & 51.8 \footnotesize{$\pm$ 4.1} & 40.4 \footnotesize{$\pm$ 1.4}  
             & 28.9 \footnotesize{$\pm$ 0.4} & 19.5 \footnotesize{$\pm$ 1.5} & 10.5 \footnotesize{$\pm$ 0.4} & 29.8 \footnotesize{$\pm$ 0.9} & 20.1 \footnotesize{$\pm$ 1.2} \\
~~ $+$ Co-teaching~\cite{han2018coteaching} & 56.1 \footnotesize{$\pm$ 2.3}& 39.8 \footnotesize{$\pm$ 4.4} & 30.5 \footnotesize{$\pm$ 3.6} & 53.5 \footnotesize{$\pm$ 2.9} & 38.7 \footnotesize{$\pm$ 1.3}  
                  & 30.4 \footnotesize{$\pm$ 0.3} & 22.0 \footnotesize{$\pm$ 1.2} & 13.3 \footnotesize{$\pm$ 0.3} & 30.8 \footnotesize{$\pm$ 1.4} & 20.3 \footnotesize{$\pm$ 0.6} \\
~~ $+$ DivideMix~\cite{li2019dividemix} & 56.1 \footnotesize{$\pm$ 1.3} & 43.7 \footnotesize{$\pm$ 0.3} & 35.1 \footnotesize{$\pm$ 1.9} & 56.1 \footnotesize{$\pm$ 0.6} & 38.9 \footnotesize{$\pm$ 2.5}
                & 30.5 \footnotesize{$\pm$ 0.7} & 23.1 \footnotesize{$\pm$ 1.6} & 13.1 \footnotesize{$\pm$ 0.5} & 30.2 \footnotesize{$\pm$ 0.5} & 20.1 \footnotesize{$\pm$ 0.7}\\

\cmidrule(lr){1-11} 

GBS~\cite{prabhu2020gdumb} & 54.8 \footnotesize{$\pm$ 1.2} & 41.8 \footnotesize{$\pm$ 0.9} & 27.3 \footnotesize{$\pm$ 2.1} & 54.2 \footnotesize{$\pm$ 1.7} & 40.4 \footnotesize{$\pm$ 1.1}  
           & 28.0 \footnotesize{$\pm$ 0.2} & 18.8 \footnotesize{$\pm$ 1.7} & 10.2 \footnotesize{$\pm$ 0.8} & 29.5 \footnotesize{$\pm$ 0.9} & 20.3 \footnotesize{$\pm$ 0.5} \\ 
~~ $+$ SELFIE~\cite{song2019selfie} & 55.4 \footnotesize{$\pm$ 0.8} & 41.6 \footnotesize{$\pm$ 0.3} & 27.8 \footnotesize{$\pm$ 1.8} & 51.3 \footnotesize{$\pm$ 4.7} & 40.7 \footnotesize{$\pm$ 0.9} 
             & 28.0 \footnotesize{$\pm$ 0.6} & 18.4 \footnotesize{$\pm$ 1.9} & 9.9 \footnotesize{$\pm$ 0.7} & 29.6 \footnotesize{$\pm$ 2.1} & 10.4 \footnotesize{$\pm$ 0.3}\\
~~ $+$ Co-teaching~\cite{han2018coteaching} & 55.1 \footnotesize{$\pm$ 0.9} & 42.7 \footnotesize{$\pm$ 0.9} & 31.4 \footnotesize{$\pm$ 3.6} & 54.0 \footnotesize{$\pm$ 0.9} & 39.5 \footnotesize{$\pm$ 1.3}
                  & 29.5 \footnotesize{$\pm$ 0.7} & 20.9 \footnotesize{$\pm$ 1.3} & 12.8 \footnotesize{$\pm$ 1.5} & 29.0 \footnotesize{$\pm$ 1.6} & 20.5 \footnotesize{$\pm$ 1.3} \\
~~ $+$ DivideMix~\cite{li2019dividemix} & 57.8 \footnotesize{$\pm$ 1.9} & 48.8 \footnotesize{$\pm$ 1.9} & 34.3 \footnotesize{$\pm$ 1.3} & 57.4 \footnotesize{$\pm$ 0.6} & 44.6 \footnotesize{$\pm$ 5.5}  
                & 29.6 \footnotesize{$\pm$ 0.8} & 21.4 \footnotesize{$\pm$ 1.1} & 13.2 \footnotesize{$\pm$ 0.2} & 28.7 \footnotesize{$\pm$ 0.7} & 19.6 \footnotesize{$\pm$ 1.1} \\

\cmidrule(lr){1-11} 

RM~\cite{bang2021rainbow} & 57.1 \footnotesize{$\pm$ 0.1} & 46.5 \footnotesize{$\pm$ 2.6} & 33.5 \footnotesize{$\pm$ 3.0} & 58.3 \footnotesize{$\pm$ 2.6} & 46.2 \footnotesize{$\pm$ 1.9} & 31.7 \footnotesize{$\pm$ 1.3} & 23.9 \footnotesize{$\pm$ 0.9} & 14.2 \footnotesize{$\pm$ 0.4} & 32.0 \footnotesize{$\pm$ 1.0} & 21.6 \footnotesize{$\pm$ 1.2} \\ 
~~ $+$ SELFIE~\cite{song2019selfie} & 56.8 \footnotesize{$\pm$ 1.3} & 44.8 \footnotesize{$\pm$ 0.6} & 31.8 \footnotesize{$\pm$ 4.4} & 57.9 \footnotesize{$\pm$ 1.5} & 46.9 \footnotesize{$\pm$ 1.9} & 32.1 \footnotesize{$\pm$ 0.8} & 22.1 \footnotesize{$\pm$ 1.0} & 12.4 \footnotesize{$\pm$ 0.7} & 31.7 \footnotesize{$\pm$ 0.4} & 20.9 \footnotesize{$\pm$ 0.9} \\
~~ $+$ Co-teaching~\cite{han2018coteaching} & 57.5 \footnotesize{$\pm$ 1.8} & 47.6 \footnotesize{$\pm$ 0.7} & 35.1 \footnotesize{$\pm$ 2.0} & 58.5 \footnotesize{$\pm$ 1.5} & 45.9 \footnotesize{$\pm$ 2.0} & 31.9 \footnotesize{$\pm$ 0.9} & 22.8 \footnotesize{$\pm$ 0.5} & 14.4 \footnotesize{$\pm$ 0.2} & 32.4 \footnotesize{$\pm$ 0.4} & 21.3 \footnotesize{$\pm$ 1.1} \\
~~ $+$ DivideMix~\cite{li2019dividemix} & 61.3 \footnotesize{$\pm$ 0.8} & 50.9 \footnotesize{$\pm$ 3.3} & 34.9 \footnotesize{$\pm$ 3.1} & 60.6 \footnotesize{$\pm$ 1.7} & 46.4 \footnotesize{$\pm$ 5.1} & 31.2 \footnotesize{$\pm$ 0.7} & 23.4 \footnotesize{$\pm$ 0.9} & 14.8 \footnotesize{$\pm$ 0.8} & 30.6 \footnotesize{$\pm$ 0.3} & 21.1 \footnotesize{$\pm$ 0.6} \\

\cmidrule(lr){1-11} 

\textbf{\our} (ours) & \textbf{61.3 \footnotesize{$\pm$ 2.1}} & \textbf{59.2 \footnotesize{$\pm$ 0.3}} & \textbf{52.4 \footnotesize{$\pm$ 2.0}} & \textbf{61.6 \footnotesize{$\pm$ 1.6}} & \textbf{47.1 \footnotesize{$\pm$ 3.2}}
              & \textbf{35.6 \footnotesize{$\pm$ 0.4}} & \textbf{33.4 \footnotesize{$\pm$ 0.3}} & \textbf{28.8 \footnotesize{$\pm$ 0.6}} & \textbf{34.6 \footnotesize{$\pm$ 0.3}} & \textbf{25.7 \footnotesize{$\pm$ 1.1}} \\
\bottomrule
\end{tabular}%
}
\vspace*{-1.5em}
\end{table*}

\vspace{-1em}\paragraph{Metrics.} 
As a main performance metric, we report the \emph{last} test\,(or validation) accuracy, which is the most widely used metric in CL literature~\cite{rwalk, han2018coteaching}, where ``last" refers to the moment when all tasks arrive. 
To quantitatively evaluate robustness in CL with noisy labels, we propose metrics of `purity' and `diversity' for the episodic memory $\mathcal{M}$ as:
\begin{equation}
\footnotesize
\label{eq:metric}
\!\begin{aligned}
& {\rm Purity} = \frac{\sum_{(x_i,\tilde{y_i}) \in \mathcal{M}} \mathbbm{1}(\tilde{y_i}=y_i)}{|\mathcal{M}|},\\
& {\rm Div.} = \frac{1}{C} \sum_{c=1}^{C} \frac{2}{|\mathcal{M}_{c}|(|\mathcal{M}_{c}|\! - \!1)} \!\!\sum_{
\substack{x_i, x_j \in \mathcal{M}_{c}\\
x_i \neq x_j}}\!\!\! ||f (x_i) \!-\! f(x_j) ||_2,\!\!\!\!\!
\end{aligned}
\end{equation}
where $f(x_i)$ is the representation vector of an example $x_i$ by a jointly trained (using all tasks, not in CL setup) model, and $\mathcal{M}_c$ is a subset of $\mathcal{M}$ consisting of training examples whose ground-truth label is $c$, and $C$ is the number of classes. 
`Purity' measures how many clean examples are correctly included in the memory. 
`Diversity' measures how much the examples in the memory are spread out in the representation space per class. We describe the trade-off of purity and diversity in the supplementary material (Sec. 4).

\vspace{-1em}\paragraph{Baselines.} 
We compare our \our with the combination of representative memory sampling methods for continual learning and robust learning methods for noisy labels. As summarized in Sec.~\ref{sec:related_work}, the memory sampling methods include \textit{reservoir sampling}\,(RSV)\,\cite{riemer2018learning}, \textit{greedy balancing sampler}\,(GBS)~\cite{prabhu2020gdumb}, and \textit{rainbow memory}\,(RM)~\cite{bang2021rainbow}. 
Since RM originally updates memory per every task, we update memory of RM at every batch for online CL scenario. %

For the robust learning baselines, we consider three widely used approaches in the literature for addressing noisy labels; a re-labeling approach \textit{SELFIE}~\cite{song2019selfie}, a sample selection approach \textit{Co-teaching}~\cite{han2018coteaching}, and a semi-supervised learning approach \textit{DivideMix}~\cite{li2019dividemix}. 
We compared \our to them; three memory only sampling methods with three robust learning methods combined plus three using memory sampling only, total of twelve.

\vspace{-1em}
\paragraph{Implementation Details.} We use ResNet~\cite{he2016resnet} family as a backbone for all compared algorithms; ResNet18, ResNet32, ResNet34 and ResNet34 are used for CIFAR-10/100, Food-101N, and WebVision, respectively. 
For CIFARs, we train ResNet using an initial learning rate of $0.05$ with cosine annealing and a batch size of 16 for 256 epochs, following \cite{bang2021rainbow}.
For Food-101N and WebVision, we use the same training hyper-parameters for CIFARs except the training epochs (256 $\to$ 128). 
In all experiments, we use 1.0 for $\eta$ in Eq.~\ref{eq:final}, and use CutMix~\cite{yun2019cutmix} and AutoAugment~\cite{cubuk2019autoaugment} for data augmentation with the memory.
For reliable validation, we repeat all the experiments thrice with a single NVIDIA P40 GPU and report the average results. 

{
\newcolumntype{L}[1]{>{\raggedright\let\newline\\\arraybackslash\hspace{0pt}}m{#1}}
\newcolumntype{X}[1]{>{\centering\let\newline\\\arraybackslash\hspace{0pt}}p{#1}}
\begin{table}[t]
\caption{Last validation accuracy on WebVision (K=1,000) and Food-101N (K=2,000), where $K$ is episodic memory size.}
\vspace{-2em}
\begin{center}
\footnotesize
\begin{tabular}{L{3.6cm} L{1.75cm} X{1.75cm}}
\toprule
Methods & WebVision & Food-101N \\\midrule
RSV~\cite{riemer2018learning} & 18.4 \footnotesize{$\pm$ 0.8} & 11.6 \footnotesize{$\pm$ 1.0} \\
RSV$+$ SELFIE~\cite{song2019selfie}& 19.1 \footnotesize{$\pm$ 0.4} & 11.1 \footnotesize{$\pm$ 1.3} \\
RSV$+$ Co-teaching~\cite{han2018coteaching}& 16.5 \footnotesize{$\pm$ 0.3} & 11.3 \footnotesize{$\pm$ 1.0} \\ 
RSV$+$ DivideMix~\cite{li2019dividemix} & 11.7 \footnotesize{$\pm$ 1.1} & 7.3 \footnotesize{$\pm$ 0.4} \\

\cmidrule(lr){1-3} 

GBS~\cite{prabhu2020gdumb} & 20.3 \footnotesize{$\pm$ 2.1} & 10.5 \footnotesize{$\pm$ 0.4} \\
GBS $+$ SELFIE~\cite{song2019selfie} & 20.0 \footnotesize{$\pm$ 2.1} & 10.4 \footnotesize{$\pm$ 0.3} \\
GBS $+$ Co-teaching~\cite{han2018coteaching} & 17.8 \footnotesize{$\pm$ 1.8} & 10.5 \footnotesize{$\pm$ 0.2} \\
GBS $+$ DivideMix~\cite{li2019dividemix} & 14.0 \footnotesize{$\pm$ 0.6} & 6.9 \footnotesize{$\pm$ 0.1} \\

\cmidrule(lr){1-3} 

RM~\cite{bang2021rainbow} & 23.1 \footnotesize{$\pm$ 0.1} & 11.7 \footnotesize{$\pm$ 0.2} \\
RM$+$ SELFIE~\cite{song2019selfie} & 21.4 \footnotesize{$\pm$ 1.1} & 12.0 \footnotesize{$\pm$ 0.2} \\
RM$+$ Co-teaching~\cite{han2018coteaching} & 21.8 \footnotesize{$\pm$ 0.8} & 12.0 \footnotesize{$\pm$ 0.4} \\ 
RM$+$ DivideMix~\cite{li2019dividemix} & 18.2 \footnotesize{$\pm$ 0.8} & 9.2 \footnotesize{$\pm$ 0.9} \\\midrule
\textbf{\our} (Ours) & \textbf{25.8 \footnotesize{$\pm$ 0.7}}  & \textbf{13.4 \footnotesize{$\pm$ 0.3}} \\ \bottomrule
\end{tabular}
\end{center}
\label{tab:real_result}
\vspace{-1em}
\end{table}
}

\begin{comment}
{
\begin{table}[t]

\centering
\caption{Last validation accuracy on WebVision (K=1,000) and Food-101N (K=2,000), where $K$ is episodic memory size.}
\vspace{-1.5em}
\label{tab:real_result}
\resizebox{0.85\linewidth}{!}{%

\begin{tabular}{@{}lrr@{}}
%
\toprule 
Methods & WebVision & Food-101N \\ 

\cmidrule(lr){1-3} 

%
%
%
%
RSV~\cite{riemer2018learning} & 18.4 \footnotesize{$\pm$ 0.8} & 11.6 \footnotesize{$\pm$ 1.0} \\
RSV$+$ SELFIE~\cite{song2019selfie}& 19.1 \footnotesize{$\pm$ 0.4} & 11.1 \footnotesize{$\pm$ 1.3} \\
RSV$+$ Co-teaching~\cite{han2018coteaching} & 16.5 \footnotesize{$\pm$ 0.3} & 11.3 \footnotesize{$\pm$ 1.0} \\ 
RSV$+$ DivideMix~\cite{li2019dividemix} & 11.7 \footnotesize{$\pm$ 1.1} & 7.3 \footnotesize{$\pm$ 0.4} \\

\cmidrule(lr){1-3} 

%
%
%
%
GBS~\cite{prabhu2020gdumb} & 20.3 \footnotesize{$\pm$ 2.1} & 10.5 \footnotesize{$\pm$ 0.4} \\
GBS $+$ SELFIE~\cite{song2019selfie} & 20.0 \footnotesize{$\pm$ 2.1} & 10.4 \footnotesize{$\pm$ 0.3} \\
GBS $+$ Co-teaching~\cite{han2018coteaching} & 17.8 \footnotesize{$\pm$ 1.8} & 10.5 \footnotesize{$\pm$ 0.2} \\
GBS $+$ DivideMix~\cite{li2019dividemix} & 14.0 \footnotesize{$\pm$ 0.6} & 6.9 \footnotesize{$\pm$ 0.1} \\

\cmidrule(lr){1-3} 
\textbf{\our} (ours) & \textbf{25.8 \footnotesize{$\pm$ 0.7}}  & \textbf{13.4 \footnotesize{$\pm$ 0.3}}  \\

\bottomrule
\end{tabular}%
%
%
}
\end{table}
}
\end{comment}

\begin{figure*}[t!]
    \centering
    \begin{subfigure}[t]{0.495\linewidth}
        \includegraphics[width=0.49\textwidth]{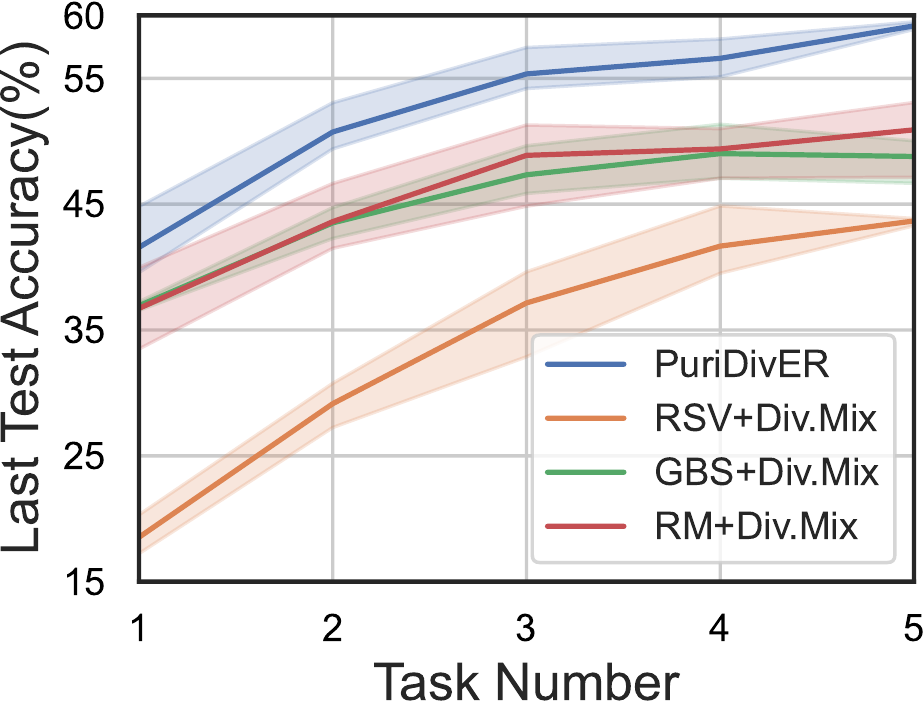}
        \hfill
        \includegraphics[width=0.496\textwidth]{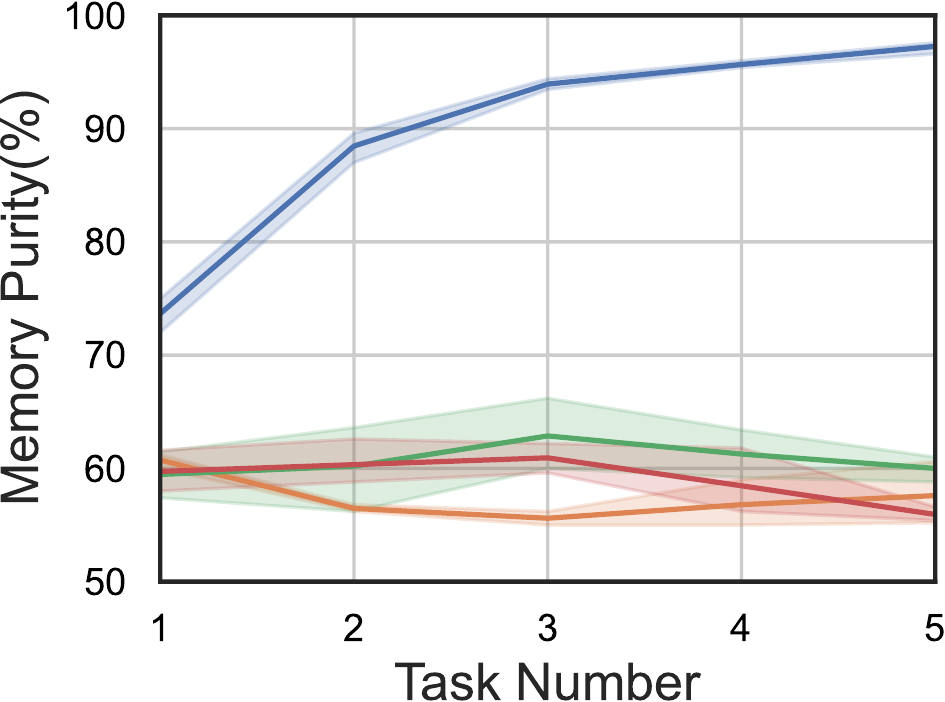}
        \caption{CIFAR-10}
        \label{fig:cifar10_sym40}
    \end{subfigure}
    \begin{subfigure}[t]{0.495\linewidth}
        \includegraphics[width=0.49\textwidth]{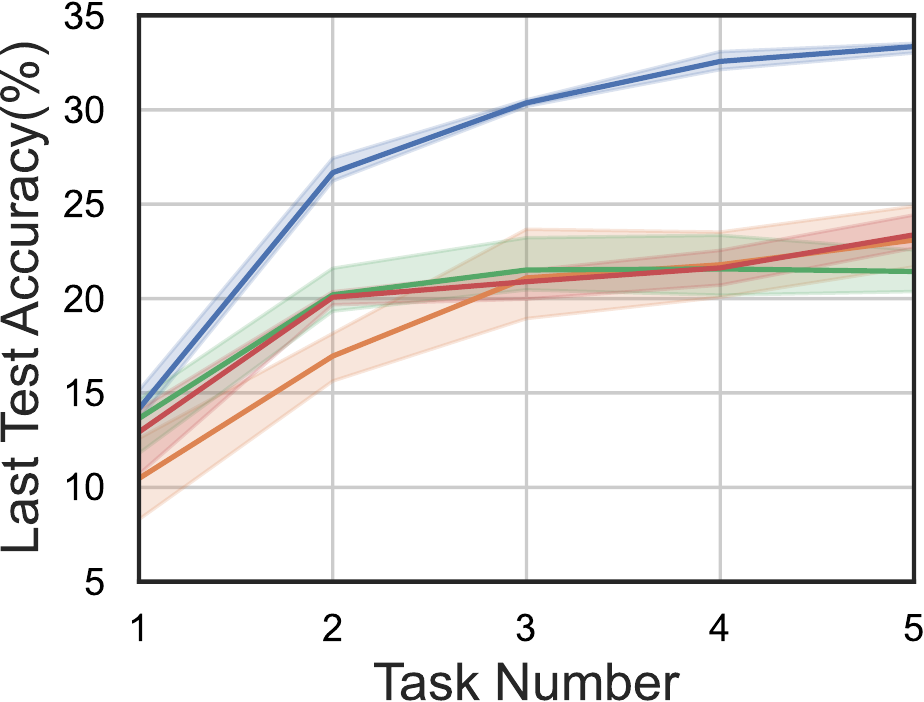}
        \hfill
        \includegraphics[width=0.496\textwidth]{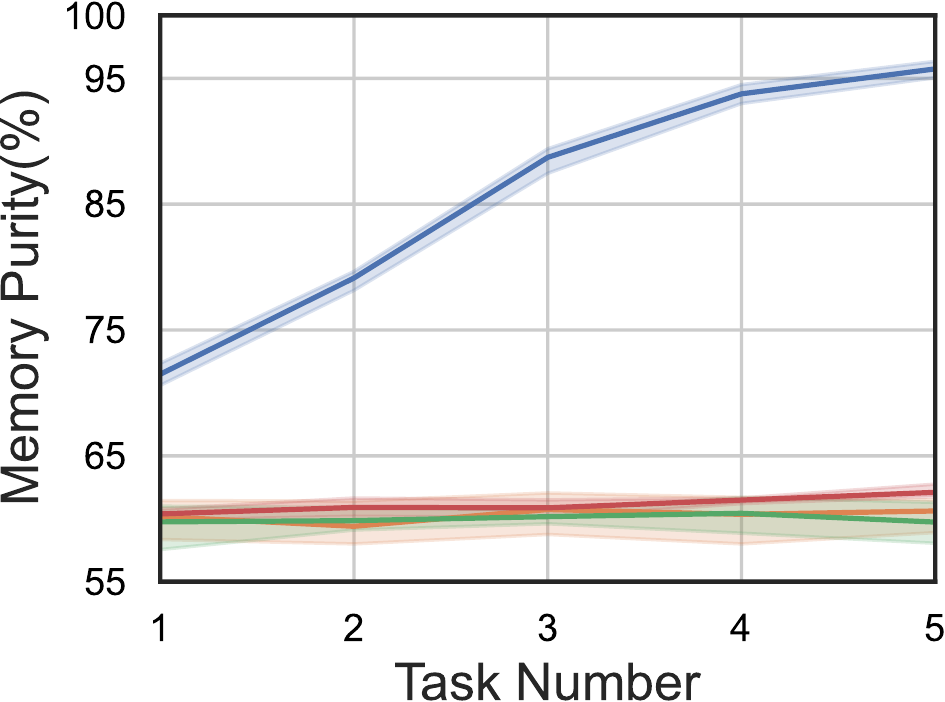}
        \caption{CIFAR-100}
        \label{fig:cifar100_sym40}
    \end{subfigure}
    \vspace{-1em}
    \caption{Last test accuracy as a function of memory purity when the task injection progresses on CIFAR-10/100 with SYM-40\%. We provide more results with various noise ratio in the supplementary material (Sec. 3).}
    \label{fig:cifar_result}
    \vspace{-1em}
\end{figure*}

\begin{figure}[t!]
    \centering
    \begin{subfigure}[t]{0.49\linewidth}
        \includegraphics[width=\textwidth]{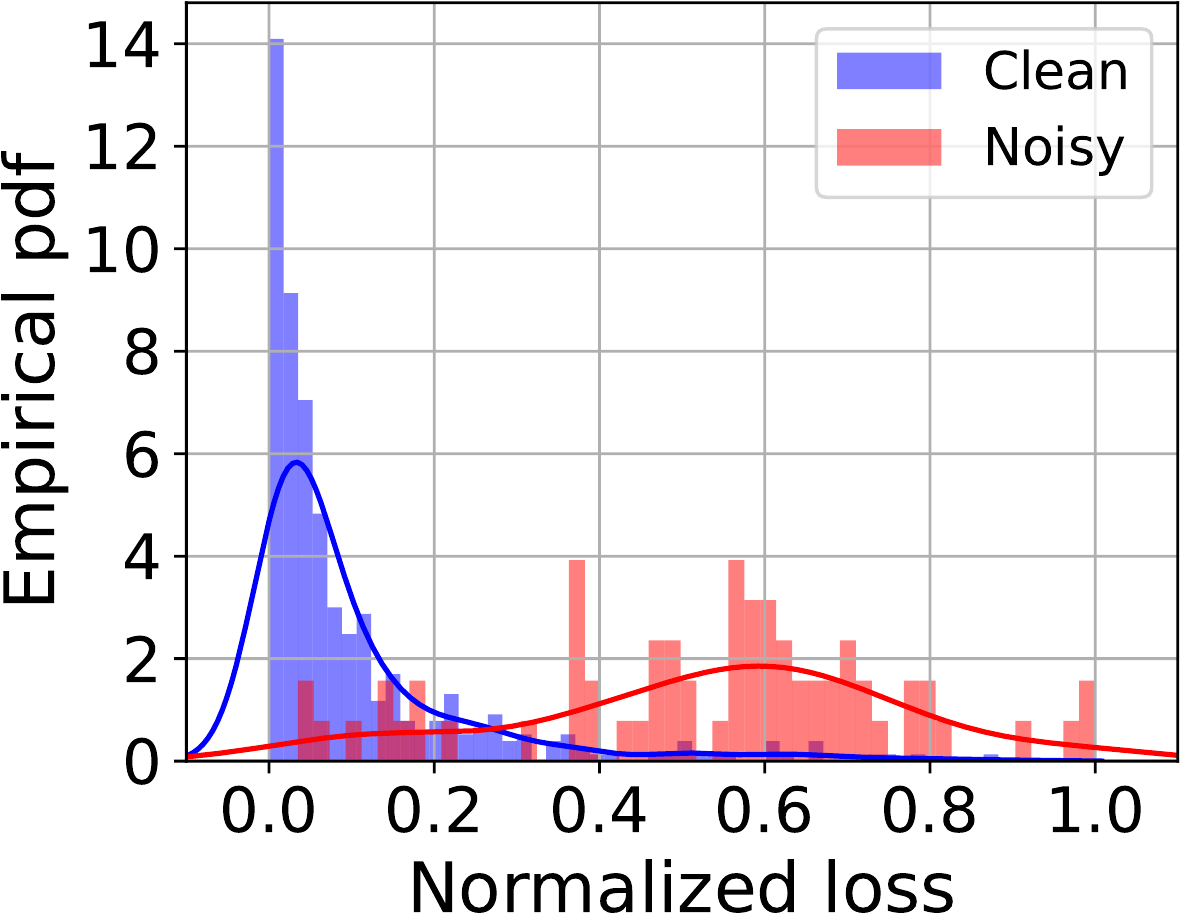}
        \label{fig:historgram_loss}
    \end{subfigure}
    \begin{subfigure}[t]{0.49\linewidth}
        \includegraphics[width=\textwidth]{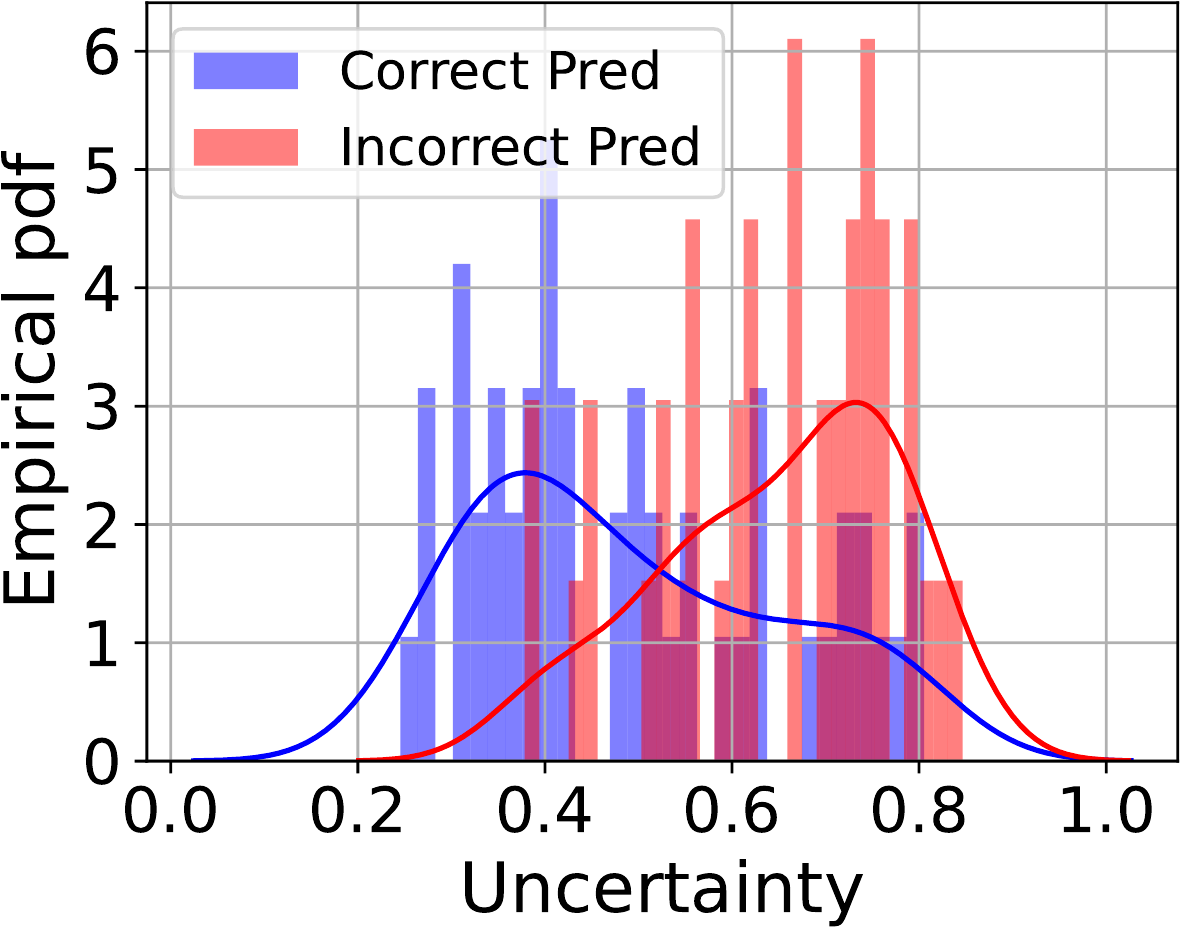}
        \label{fig:histogram_uncert}
    \end{subfigure}
    \vspace{-1.8em}
    \caption{Distribution of examples in the episodic memory training on CIFAR-10 with SYM-20\%. GMMs effectively distinguish between (left) clean and noisy labels by loss, and (right) correct and incorrect predictions by uncertainty (Eq.~\ref{eq:uncertainty})}
    \label{fig:validity_gmm_fitting}
\end{figure}

\subsection{Results}
\label{sec:results}

We compare the proposed \our with twelve baselines on the two CIFARs datasets with varying symmetric and asymmetric noise, and summarize the results in Tab.~\ref{tab:cifar_result}. 
We also compare methods on real-world noisy datasets; WebVision and Food-101N datasets in Tab.~\ref{tab:real_result}.

As shown in the two tables, \our consistently outperforms all other compared methods. 
Comparing \our with other methods, in particular, accuracy increases up to 16\% on CIFARs, 11\% and 3\% on WebVision and Food-101N, respectively.
Interestingly, the performance gain becomes large as the noise ratio increases or the difficulty of training data increases from CIFAR-10 to 100.

Noticeably, SELFIE~\cite{song2019selfie} does not improve the accuracy in average when combined with the three memory sampling approaches. 
We believe that the SELFIE rather produces many false corrections in the online learning setup, since it requires an abundant number of data for re-labeling unlike episodic memory that contains a small number of data.

Despite the use of multiple networks, Co-teaching~\cite{han2018coteaching} marginally improves the performance. %
Unlike SELFIE and Co-teaching, DivideMix considerably improves the accuracy in CIFAR-10, but not in CIFAR-100 with asymmetric noise and real noise datasets, which have more realistic label noise. 
We expect that the size of (expected) noisy dataset is larger when splitting whole dataset into the clean and the noisy data, as noise ratio becomes larger or learning becomes more difficult. 
As such, it has too small (expected) clean data to train the model due to the small size of episodic memory. In Tab.~\ref{tab:real_result}, we observe that \our presents a competitive capability on noisy CL tasks in real-world label corruption while outperforming comparable methods.

\subsection{Detailed Analyses}
\label{sec:ssl_sr}

\subsubsection{Purity of Episodic Memory}
We first investigate the effect of memory purity on the performance of our method.
We measure the last test accuracy as a function of the memory purity (Eq.~\ref{eq:metric}). %
We compare our \our with RSV, GBS and RM integrated with DivideMix in Fig.~\ref{fig:cifar_result}. 
As shown in the figure, ours significantly outperforms them in both accuracy and memory purity on CIFAR-10 and CIFAR-100 by large margins. 

Particularly, the performance gap between \our and other methods gradually increase as more tasks arrive by an synergistic effect of our episodic memory construction\,(Sec.~\ref{sub:memory_update}) and memory usage schemes\,(Sec.~\ref{sec:sub:robustlearning}). 
Interestingly, the performance gain of \our increases as data becomes more difficult. It implies that the purity of other baselines relatively much worsen in CIFAR-100, while ours maintains its significant dominance. There are similar gains on various noise ratio as shown in the supplementary material (Sec. 3).

\vspace{-1em}
\subsubsection{Ablations on Robust Learning Components}
We now investigate the contribution of each component of the proposed robust learning approach to the accuracy. %
The both sides of Fig.~\ref{fig:validity_gmm_fitting} show the loss distribution of all examples in the memory $\mathcal{M}$ and the uncertainty distribution (Eq.~\ref{eq:uncertainty_loss}) of the examples in (expected) noisy set $\mathcal{N}$, respectively.
The loss distributions in Fig.~\ref{fig:validity_gmm_fitting}-(left) are bi-modal for clean and noisy examples, thus clearly separating them by fitting GMMs. 
Likewise, the uncertainty distributions in Fig.~\ref{fig:validity_gmm_fitting}-(right) are bi-modal as well for the correctness of model's predictions. 
Hence, re-labeling using the model's prediction for the left Gaussian components (\ie, certain examples) ensures that the proposed re-labeling method achieves high precision.

\begin{table}[t]
\centering
\caption{Ablation study for \our on CIFAR-10 and CIFAR-100 with various noise ratios and types. SYM and ASYM refer to the symmetric and asymmetric label noise, respectively. }
\vspace{-0.5em}
\label{tab:dpm_ablation}
\resizebox{1.0\linewidth}{!}{%
\begin{tabular}{@{}ccrrrrrrrrrr@{}}
\toprule 
& & \multicolumn{5}{c}{CIFAR-10 (K=500)} & \multicolumn{5}{c}{CIFAR-100 (K=2,000)} \\ 
\multicolumn{2}{c}{Robust Learning}& \multicolumn{3}{c}{SYM} & \multicolumn{2}{c}{ASYM} & \multicolumn{3}{c}{SYM} & \multicolumn{2}{c}{ASYM} \\ 
Re-label & Consistency &
\multicolumn{1}{c}{20} & \multicolumn{1}{c}{40} & \multicolumn{1}{c}{60} & \multicolumn{1}{c}{20} & \multicolumn{1}{c}{40} & 
\multicolumn{1}{c}{20} & \multicolumn{1}{c}{40} & \multicolumn{1}{c}{60} & \multicolumn{1}{c}{20} & \multicolumn{1}{c}{40} \\

\cmidrule(lr){1-2} \cmidrule(lr){3-5} \cmidrule(lr){6-7} \cmidrule(lr){8-10} \cmidrule(lr){11-12} 

&  & \textbf{61.8} & 55.4 & 46.9 & 60.6 & 46.4  
& 33.1 & 26.8 & 18.6 & 31.5 & 21.3 \\ 
\checkmark &  & 57.7 & 55.4 & 44.1 & 61.0 & 46.6  
& 34.7 & 31.5 & 26.2 & 33.2 & 22.7 \\ 
 & \checkmark & 48.6 & 46.2 & 31.3 & 52.1 & 36.7  
& 31.7 & 24.3 & 18.1 & 29.9 & 20.3 \\ 

\cmidrule(lr){1-12}
\checkmark & \checkmark & 61.3 & \textbf{59.2} & \textbf{52.4} & \textbf{61.6} & \textbf{47.1}
                        & \textbf{35.6} & \textbf{33.4} & \textbf{28.8} & \textbf{34.6} & \textbf{25.7} \\ 
\bottomrule
\end{tabular}%
}
\vspace{-0.5em}
\end{table}

Tab.~\ref{tab:dpm_ablation} summarizes the contribution of the re-labeling and consistency regularization in \our. 
Using either re-labeling or consistency regularization alone does not improve the accuracy in many experiments because re-labeling would make many false correction, and consistency regularization utilizes only a small size of clean set $\mathcal{C}$. 
For those reasons, they also drastically decrease the performance as the noise ratio increases.

In contrast, training the model for the clean and re-labeled set $\mathcal{C} \cup \mathcal{R}$ with the consistency loss for the unlabeled set $\mathcal{U}$ significantly improves the accuracy by up to 10.2\% compared to the baseline without them. 
By using the both methods, since a good number of examples in the re-label set $\mathcal{R}$ would be correctly labeled, the $\mathcal{R}$ can also be utilized to train the model better with clean set $\mathcal{C}$. 
Further, as we re-split the three sets $\mathcal{C}$, $\mathcal{R}$ and $\mathcal{U}$ at every epoch, it is likely that many data will have both effects of re-labeling and consistency regularization over the epochs, which leads to further improvement.

\begin{table}[t!]
\centering
\caption{Last test accuracy over various static $\alpha$ on CIFAR-10 and CIFAR-100 with SYM-\{20\%, 40\%, 60\%\} and ASYM-\{20\%, 40\%\}. $K$ is the size of episodic memory.} 
\vspace{-1em}
\label{tab:various_coeff}
\resizebox{1.0\linewidth}{!}{%
\begin{tabular}{@{}crrrrrrrrrr@{}}
\toprule 
& \multicolumn{5}{c}{CIFAR-10 (K=500)} & \multicolumn{5}{c}{CIFAR-100 (K=2,000)} \\
\cmidrule(lr){2-6} \cmidrule(lr){7-11} 
& \multicolumn{3}{c}{Sym.} & \multicolumn{2}{c}{Asym.} & \multicolumn{3}{c}{Sym.} & \multicolumn{2}{c}{Asym.} \\ 
$\alpha$ & 
\multicolumn{1}{c}{20} & \multicolumn{1}{c}{40} & \multicolumn{1}{c}{60} & \multicolumn{1}{c}{20} & \multicolumn{1}{c}{40} & 
\multicolumn{1}{c}{20} & \multicolumn{1}{c}{40} & \multicolumn{1}{c}{60} & \multicolumn{1}{c}{20} & \multicolumn{1}{c}{40} \\

\cmidrule(lr){1-1} \cmidrule(lr){2-4} \cmidrule(lr){5-6} \cmidrule(lr){7-9} \cmidrule(lr){10-11} 
0.1 & 57.4 & 55.1 & 51.6 & 58.9 & 47.4
    & 36.1 & 33.6 & \textbf{28.1} & 34.0 & \textbf{25.6} \\ 

0.3 & 58.1 & 55.3 & \textbf{52.6} & 58.4 & \textbf{49.0}
    & 35.5 & \textbf{33.6} & 27.3 & 34.6 & 25.5 \\
0.5 & \textbf{60.5} & \textbf{57.0} & 47.1 & \textbf{60.4} & 46.7
    & \textbf{37.0} & 32.3 & 23.0 & \textbf{36.6} & 25.5 \\ 
    
0.6 & 29.5 & 30.3 & 24.4 & 36.0 & 33.7
    & 25.3 & 18.9 & 13.5 & 26.7 & 18.2 \\ 
    
1.0 & 26.1 & 19.4 & 19.3 & 33.0 & 28.1
    & 15.2 & 10.5 & 7.2 & 20.7 & 16.6 \\ 
\bottomrule
\end{tabular}
}
\vspace{-1.0em}
\end{table}

\begin{figure}[t!]
    \centering
    \begin{subfigure}[t]{0.49\linewidth}
        \includegraphics[width=\textwidth]{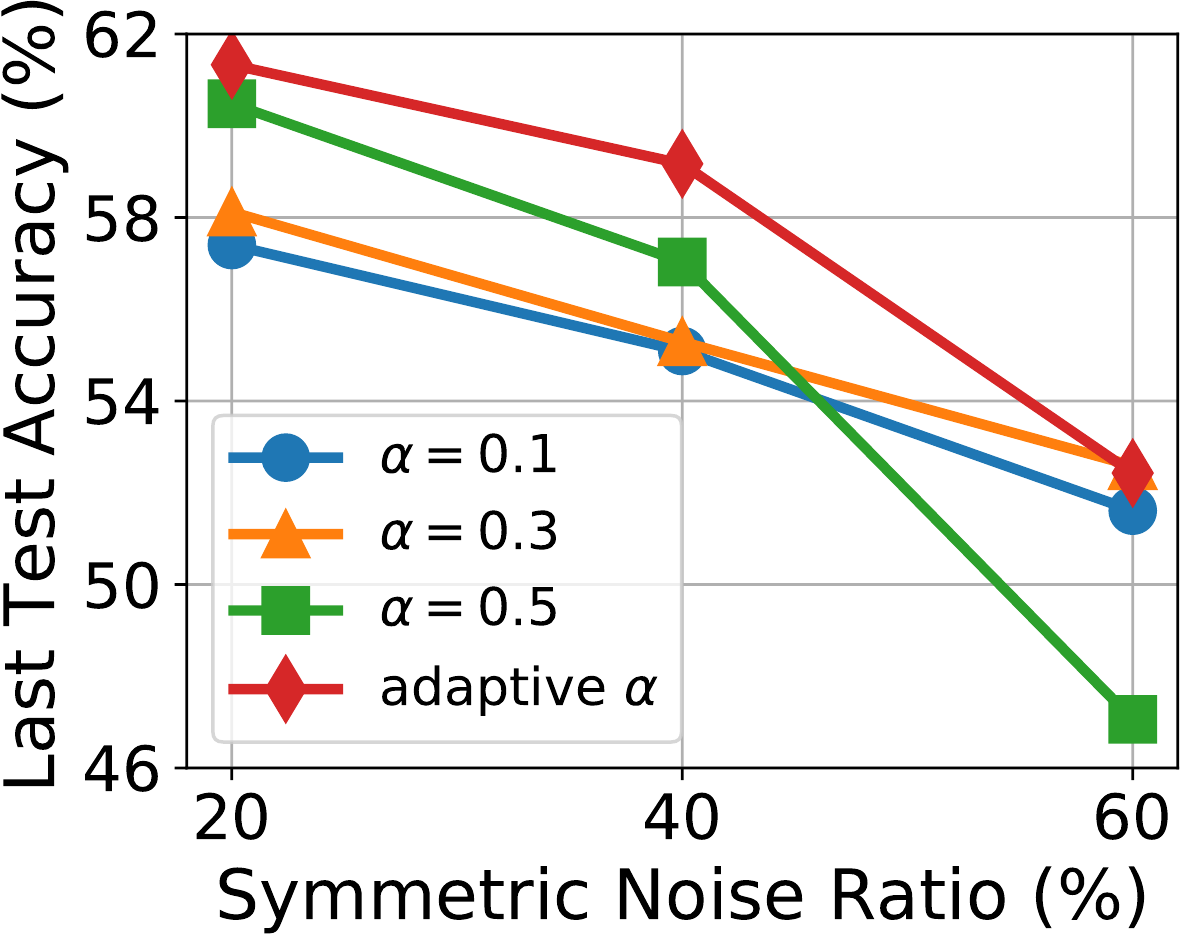}
        \label{fig:various_coeff_a}
    \end{subfigure}
    \begin{subfigure}[t]{0.49\linewidth}
        \includegraphics[width=\textwidth]{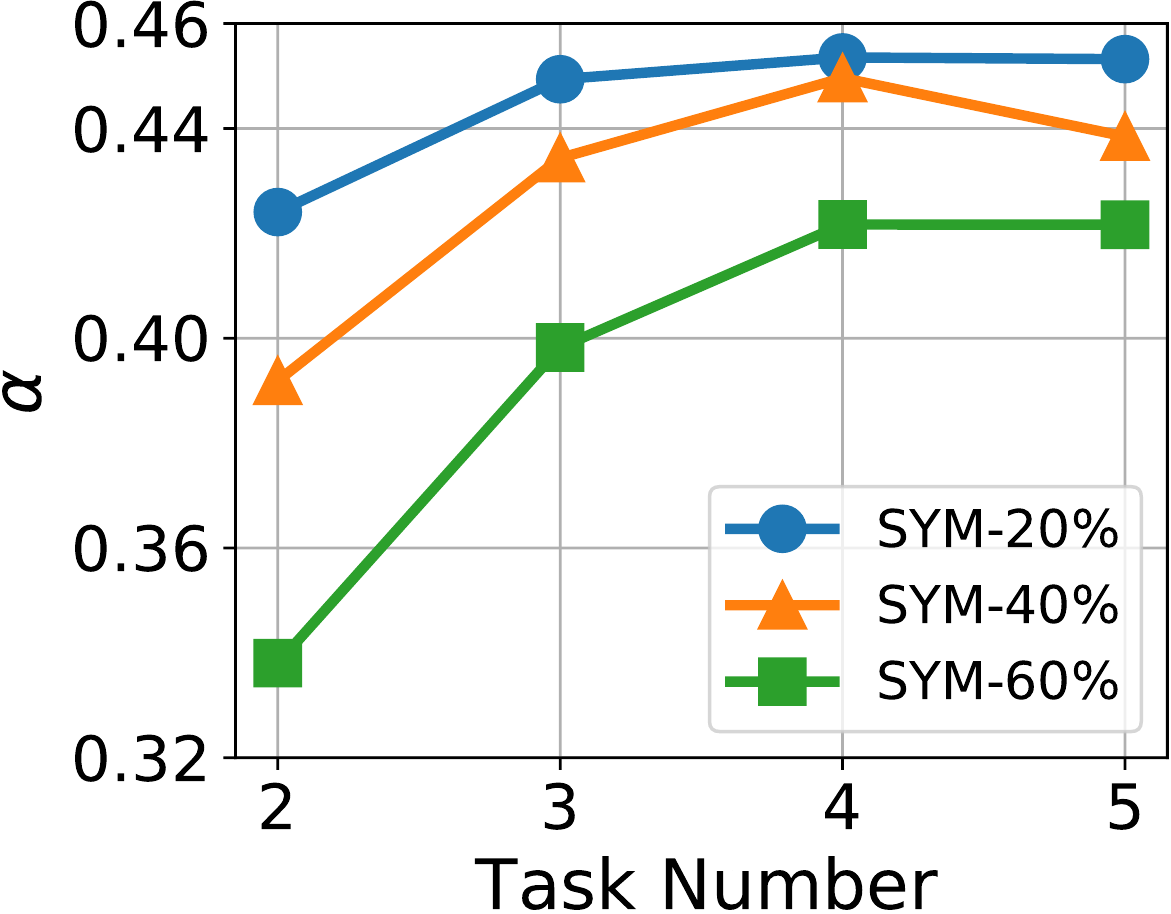}
        \label{fig:various_coeff_b}
    \end{subfigure}
    \vspace{-2em}
    \caption{Accuracy by different $\alpha$ on CIFAR-10 with SYM-\{20\%, 40\%, 60\%\}; (left) comparison of the last test accuracy with and without the adaptive strategy, (right) comparison of the mean of $\alpha$ for each task with the adaptive strategy.}
    \vspace{-0.5em}
    \label{fig:various_coeff}
\end{figure}

\vspace{-0.5em}
\subsubsection{Balancing Coefficient}
\label{sec:coefficent}
Tab.~\ref{tab:various_coeff} illustrates the effect of balancing coefficient $\alpha$ on the accuracy. 
As shown in the table, the test accuracy decreases drastically when $\alpha > 0.5$ because diversity dominantly influences memory sampling rather than memory purity. 
In the presence of label noise, this suggests that memory purity plays a significant role in performance, as opposed to the conventional CL methods that mainly emphasize example diversity~\cite{bang2021rainbow}. 
Hence, we set the range of adaptive balancing coefficient $\alpha$ to (0, 0.5) instead of (0, 1).
Additionally note that the best $\alpha$ value depends on the noise ratio. %
Generally, with increased noise ratios, a smaller $\alpha$ incurs higher accuracy since memory purity is far more important than diversity when label noise is severe. 

\vspace{-1em}\paragraph{Efficacy of Adaptive Strategy.}
We compare the performance of \our with and without using the proposed adaptive balancing coefficient strategy, and summarize the results in Fig.~\ref{fig:various_coeff}-(left). 
\our with the adaptive strategy always outperforms others. 
In addition, as shown in Fig.~\ref{fig:various_coeff}-(right), the $\alpha$ values are properly determined by \our with the adaptive strategy for three difference noise scenarios. 
The $\alpha$ tends to increase as the task number increases except for the initial task. 
It implies that small-loss\,(clean) examples are favored at the earlier stage of training for robust learning, while diverse examples are favored at the later stage of training for better generalization. 
Therefore, the proposed adaptive strategy allows \our to exploit the best $\alpha$ value for the various noise scenario.
\vspace{-1em}

\setlength{\columnsep}{10pt}%
\begin{wrapfigure}{r}{0.4\linewidth}
    \vspace{-2em}
    \includegraphics[width=1\linewidth]{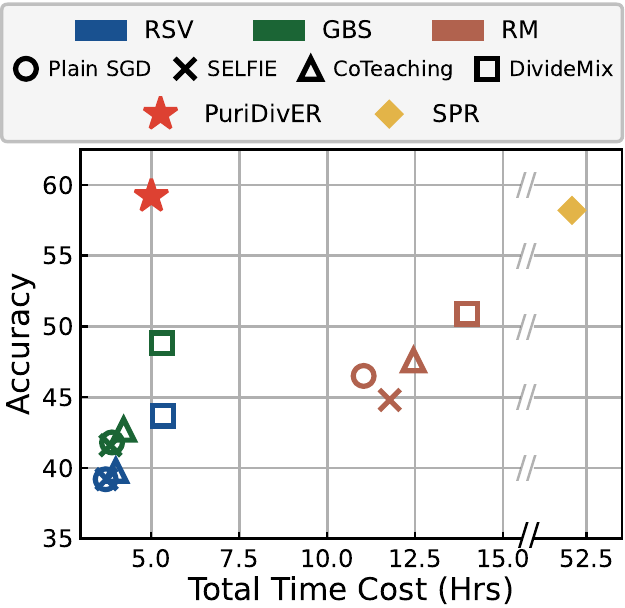}
    \vspace{-2em}
    \caption{Total training time and accuracy on CIFAR-10 SYM-40\% for various methods.}
    \label{fig:time}
    \vspace{-1.5em}
\end{wrapfigure}

\subsubsection{Time Cost}
In online continual learning, time cost is significant for the model to adapt the distribution change of data stream. Unfortunately, there is a trade-of between time and accuracy. In Fig.~\ref{fig:time}, RSV, GBS and RM have small amount of training time but poor accuracy, while SPR takes too much time to be utilized in online setup. 
Despite the additional time cost for memory construction compared to RSV and GBS, \our exhibits a good trade-off between the accuracy and time cost, compared to others.

\section{Conclusion}

We address a setup of online and blurry CL with noisy labels, which occurs frequently in real world AI deployment scenarios.
Specifically, we propose a method to address this practical task by adaptively balancing diversity and purity by an episodic memory management scheme followed by a robust learning with semi-supervised learning for additional diversity and purity enforcement for the memory usage. 
We also define an adaptively balancing coefficient to accommodate various noise ratio and sample-wise learning efficiency. 
We empirically validate that our \our not only improves accuracy over combinations of prior arts but also helps better sampling for memory.

\vspace{-1.3em}\paragraph{Limitation.}
Although \our outperforms prior arts by large margins, developing methods in noisy CL setups are in its nascent steps; 
\eg, no new class is added as the task progresses. 
Also, scaled-up empirical validation is important and desirable.

\vspace{-1.3em}\paragraph{Potential Negative Societal Impact.} 
As the CL method is for deploying AI to broader context, the data can come from any sources thus the data privacy issue may arise.
Although the proposed method has \emph{no intention} to allow such consequences, the method may carry such negative effects.
Efforts to prevent such concerns would be a focus of secure machine learning research.

\footnotesize{\vspace{-1.3em}\paragraph{Acknowledgement.} All authors thank NAVER Smart Machine Learning (NSML)~\cite{kim2018nsml} for experimental environments. This work was partly supported by the National Research Foundation of Korea (NRF) grant funded by the Korea government (MSIT) (No.2022R1A2C4002300) and Institute for Information \& communications Technology Promotion (IITP) grants funded by the Korea government (MSIT) (No.2020-0-01361-003 and 2019-0-01842, Artificial Intelligence Graduate School Program (Yonsei University, GIST), and No.2021-0-02068 Artificial Intelligence Innovation Hub).}

{\small
\bibliographystyle{ieee_fullname}
\bibliography{main}
}

\end{document}

\title{{\large Supplementary Material for}\\Online Continual Learning on a Contaminated Data Stream\\ with Blurry Task Boundaries}

\author{Jihwan Bang$^{1,2}$\hspace{0.5em}Hyunseo Koh$^{2,3}$\hspace{0.5em}Seulki Park$^{2,4}$\hspace{0.5em}Hwanjun Song$^{1,2}$\hspace{0.5em}Jung-Woo Ha$^{1,2}$ \hspace{0.5em}Jonghyun Choi$^{2,5,}$\thanks{corresponding author}\\
{NAVER CLOVA$^1$\hspace{1em}NAVER AI Lab$^2$\hspace{1em}GIST$^3$\hspace{1em}Seoul National Univ.$^4$\hspace{1em}Yonsei Univ.$^5$}\\
{\tt\small {\{jihwan.bang,hwanjun.song,jungwoo.ha\}@navercorp.com, hyunseo8157@gm.gist.ac.kr,}} \\
{\tt\small {seulki.park@snu.ac.kr, jc@yonsei.ac.kr}}
}

\maketitle

\noindent
\textbf{Note:} We use \bmp{blue} color to refer to figures, tables, section numbers and citations \textbf{in the main paper} (\textit{e.g.}, [\bmp{17}]). All {\color{red}red} or {\color{green}green} characters refer to figures, tables, section numbers and citations in this supplementary material.

\section{Detailed Algorithm}
\label{app:detailed}
Algorithm~\ref{alg:detailed} describes overall procedure of \our. For each task, a model is trained with online data stream $S_t$ via SGD optimizer (Lines 3--5). Since we can see the data at once by definition of online learning, episodic memory is updated every batch. To get the diversified examples in the memory, we consider the examples of memory and a new example from batch as memory candidates, then remove one that has the maximum value of score function (Lines 7--9). After constructing memory, the model trains with memory. Since the memory might have contaminated data by selecting examples considering both diversity and purity, we split the memory as three parts every epoch; clean set $\mathcal{C}$, noisy but high model's confident set $\mathcal{R}$, noisy and low model's confident set $\mathcal{U}$ (Lines 11--13). Finally, we apply to the different loss function of each set that is appropriate for handling noisy labels (Lines 14--15).

\begin{algorithm}[h!]
\small
    \caption{\ourfull}
    \label{alg:detailed}
    \begin{algorithmic}[1]
        \State {\textbf{Input:} $\mathcal{S}_{t}$: stream data at task $t$, $\mathcal{M}$: exemplars stored in a episodic memory, $T$: the number of tasks, $\theta_0$: initial model.}
	    \For {$t$ = 1 {\bf to} T}
	       \For {each mini-batch $\mathcal{B} \in \mathcal{S}_{t}$}
	       \State{/* {\color{orange}Online Training} */}
	       \State{$\theta \leftarrow \theta - \alpha \nabla \sum_{i \in \mathcal{B}} \ell(\theta(x_i), \tilde{y_i})$} 
	            \State{/* {\color{orange} Memory Construction} */}
	            \For {$x_i, \tilde{y_i}$ in $\mathcal{B}$}
	            \State{$\mathcal{M} \leftarrow \mathcal{M} \cup (x_i, \tilde{y_i})$}
	            \State{$\mathcal{M}$.remove($\arg\max_{(x,\tilde{y}) \in \mathcal{M}} S(x, \tilde{y})$) by \bmp{Eq. 2} }
	        \EndFor
	   \EndFor
       \State{/* {\color{orange} Memory Usage} */}
	   \For {$e=1$ {\bf to} MaxEpoch}
	        \State{Split $\mathcal{C}$, $\mathcal{N}$ from $\mathcal{M}$ via \bmp{Eq. 5}}
	        \State{Split $\mathcal{R}$, $\mathcal{U}$ from $\mathcal{N}$ via \bmp{Eq. 7}}
	       \For {each mini-batch $\mathcal{B_C} \in \mathcal{C}$, $\mathcal{B_R} \in \mathcal{R}$, $\mathcal{B_U} \in \mathcal{U}$ }
	        \State{$\theta \leftarrow \theta - \alpha \nabla [\ell_{cls} + \eta \ell_{reg}$]}
	       \EndFor
	   \EndFor
	\EndFor
    \end{algorithmic}
\end{algorithm}

\section{Detailed Configuration for Online Continual Learning}
\label{app:config_stream_data}

To split dataset into several tasks, we follow [\bmp{5}] to make \textit{blurry-CL}. \textit{Blurry-CL} contains two types of classes, major and minor classes. We set $L=0.1$, so the number of minor classes are 10\% of the entire of classes at each task. 
Since some labels might be incorrect, real class distribution of each task might be different. We configure CIFAR-10/100, WebVision and Food-101N as 5, 5, 10 and 5 tasks, respectively. Since Food-101N has 101 classes, the number of major classes is 21 at the last task. Furthermore, we consider online-CL which the incoming samples are presented to a model only once except for the examples from the memory.

\section{Accuracy and memory purity by various noise ratios}

\label{app:various_noise_ratio}
\begin{figure*}[h!]
    \centering
    \begin{subfigure}[t]{0.495\linewidth}
        \includegraphics[width=0.495\textwidth]{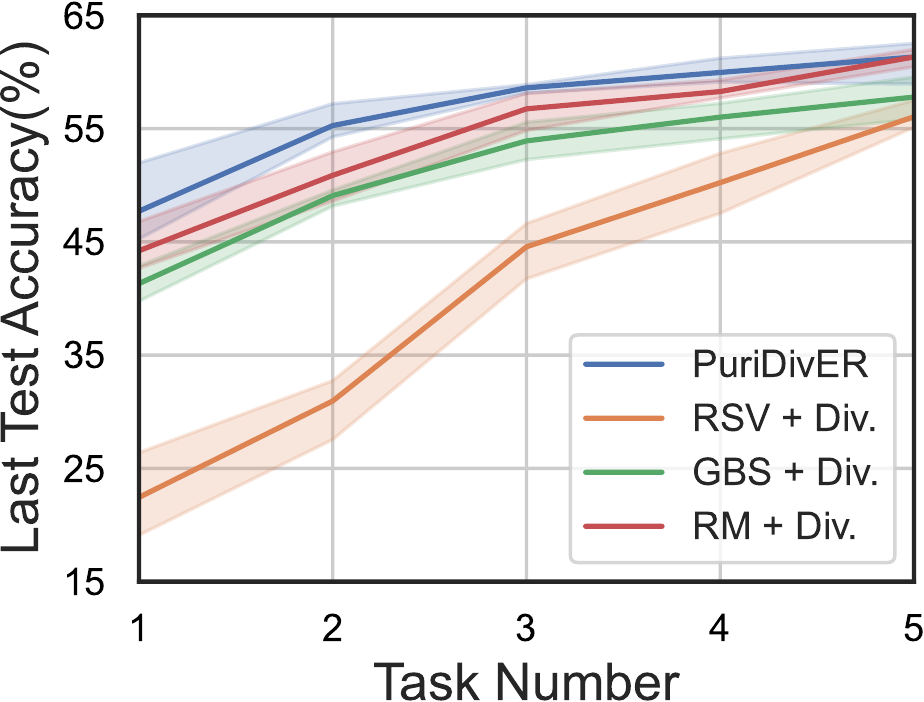}
        \hfill
        \includegraphics[width=0.495\textwidth]{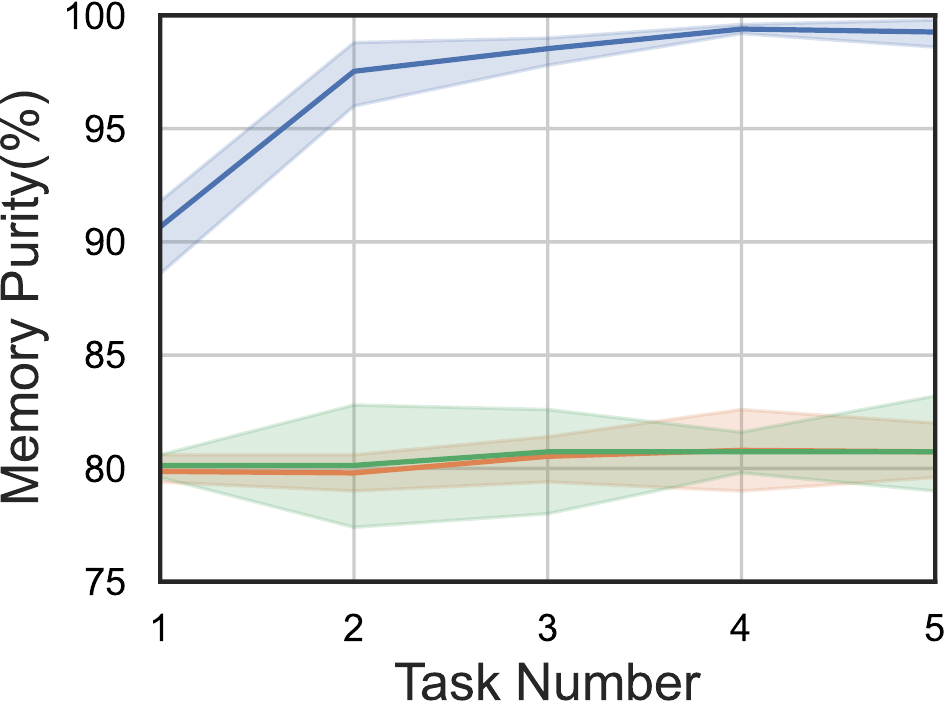}
        \caption{CIFAR-10 with SYM-20\%}
        \label{fig:cifar10_sym20}
    \end{subfigure}
    \begin{subfigure}[t]{0.495\linewidth}
        \includegraphics[width=0.495\textwidth]{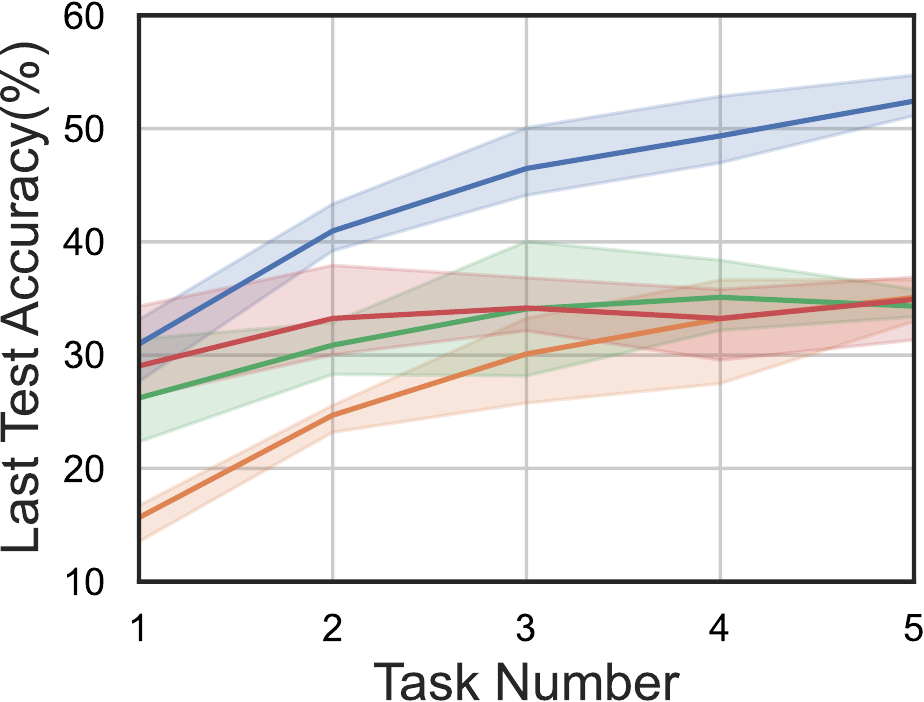}
        \hfill
        \includegraphics[width=0.495\textwidth]{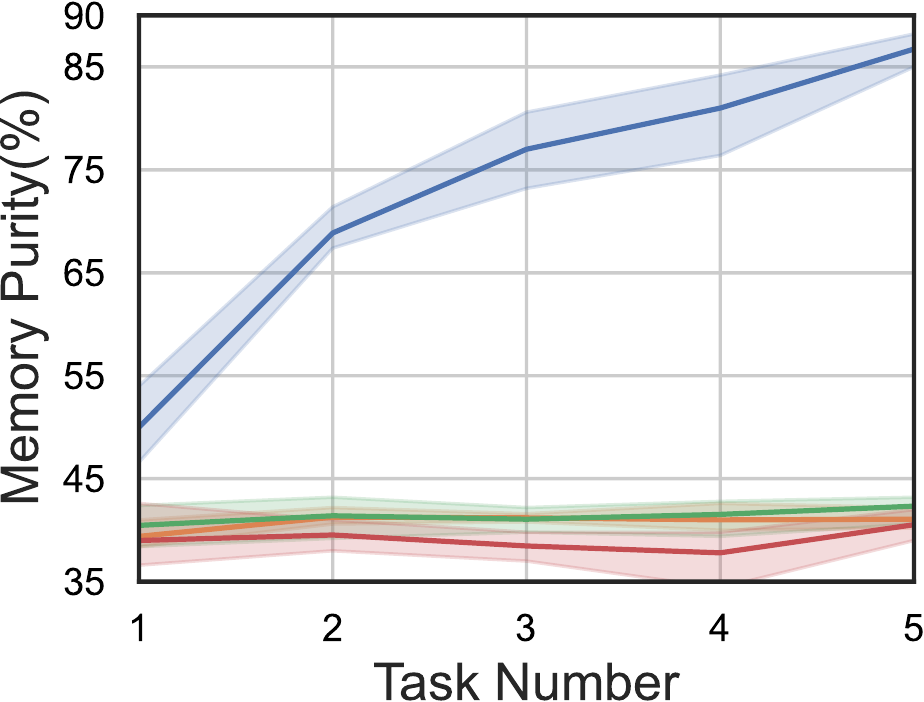}
        \caption{CIFAR-10 with SYM-60\%}
        \label{fig:cifar10_sym60}
    \end{subfigure}
    \caption{Illustration of last accuracy and memory purity changes as the task number increases on CIFAR-10 with SYM-\{20\%, 60\%\}.}
    \label{fig:cifar10_appendix_result}
\end{figure*}

\begin{figure*}[h!]
    \centering
    \begin{subfigure}[t]{0.495\linewidth}
        \includegraphics[width=0.495\textwidth]{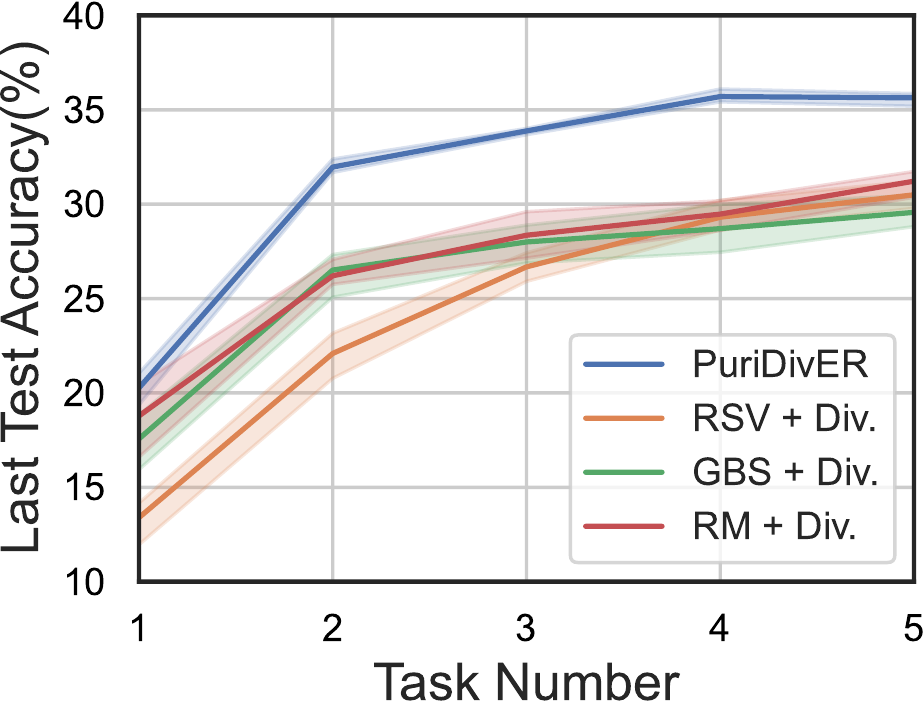}
        \hfill 
        \includegraphics[width=0.495\textwidth]{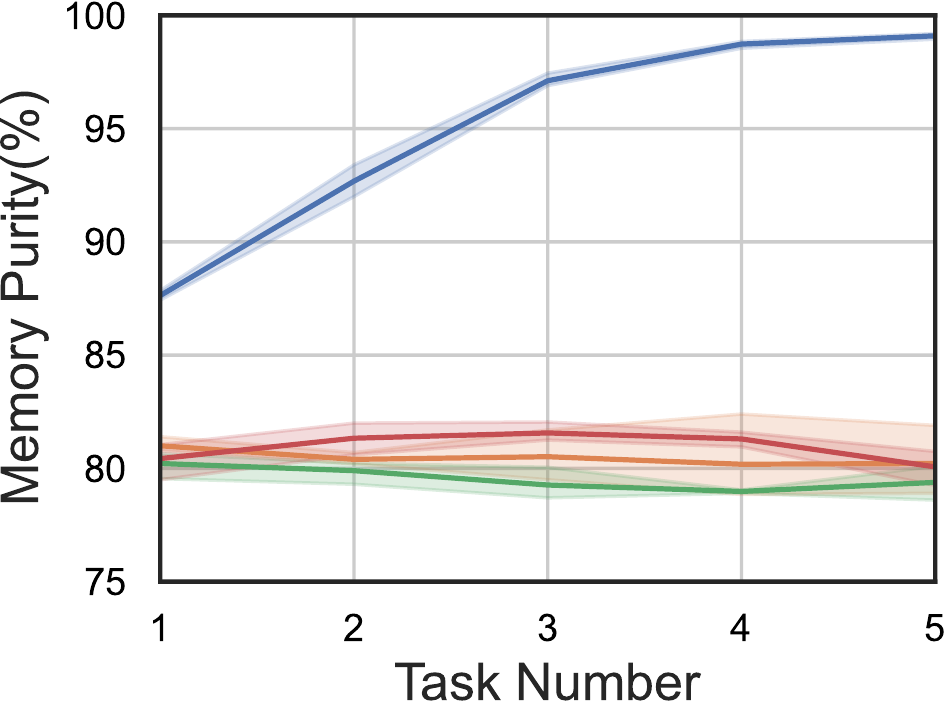}
        \caption{CIFAR-100 with SYM-20\%}
        \label{fig:cifar100_sym20}
    \end{subfigure}
    \begin{subfigure}[t]{0.495\linewidth}
        \includegraphics[width=0.495\textwidth]{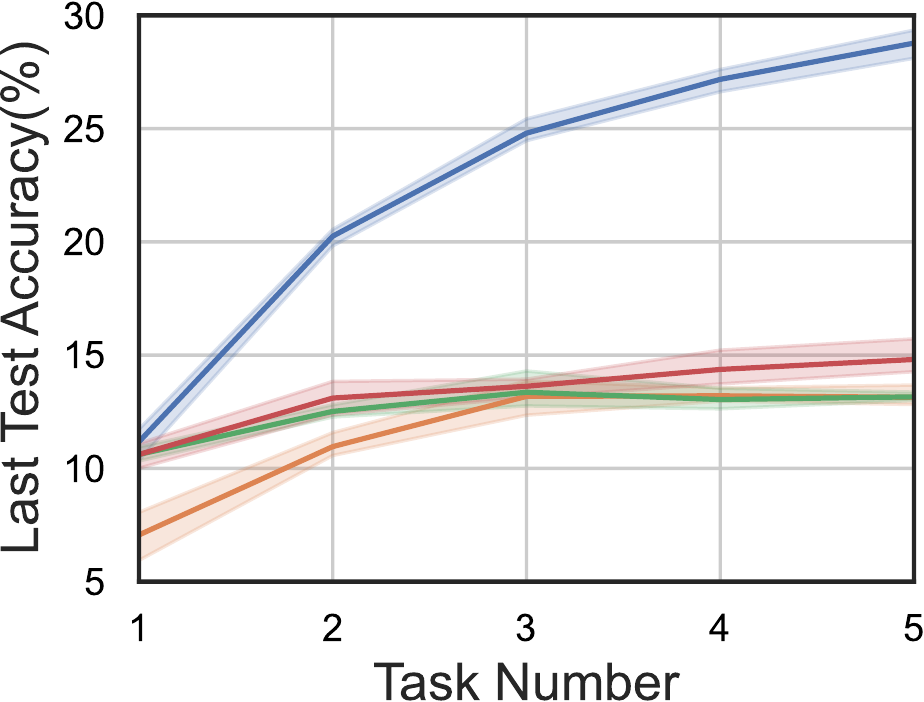}
        \hfill
        \includegraphics[width=0.495\textwidth]{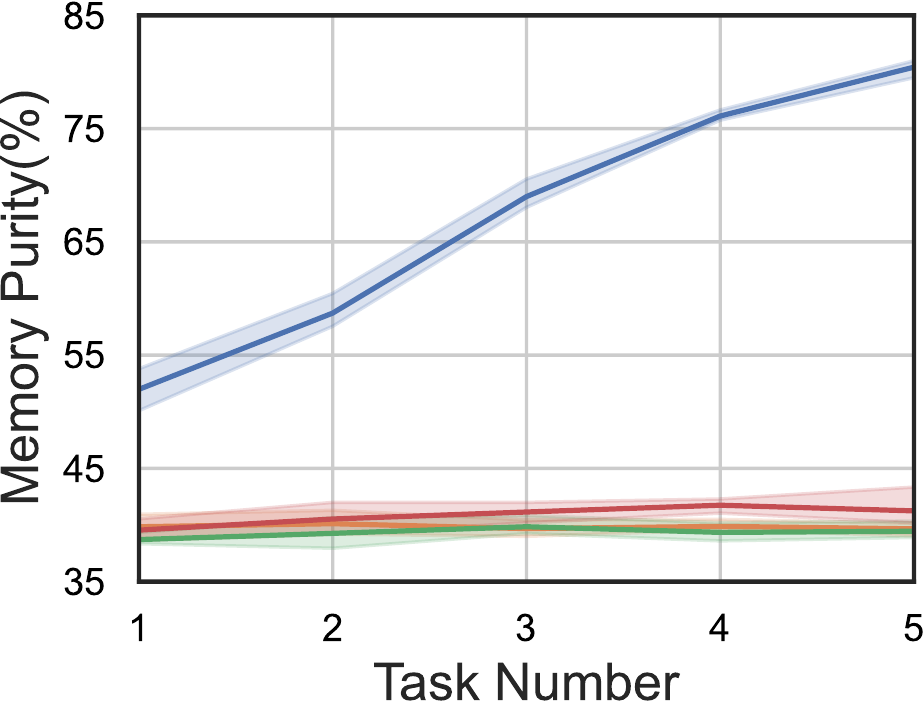}
        \caption{CIFAR-100 with SYM-60\%}
        \label{fig:cifar100_sym60}
    \end{subfigure}
    
    \caption{Illustration of last accuracy and memory purity changes as the task number increases on CIFAR-100 with SYM-\{20\%, 60\%\}.}
    \label{fig:cifar100_appendix_result}
\end{figure*}

We add experimental results of the last accuracy and memory purity of \our on CIFAR-10/100 with the symmetric noise 20\% and 60\% in Fig.~\ref{fig:cifar10_appendix_result} and ~\ref{fig:cifar100_appendix_result}.
We can find consistent results as in \bmp{Fig. 2}.
On different scenarios, \our outperforms the other baselines for both accuracy and memory purity over the entire task stream.
Especially, in memory purity, \our greatly outperforms the other baselines, and it shows that our method can effectively find clean labels from corrupted data for the online continual learning setup. Since we believe that memory sampling and robust learning from \our are the relationship to help each other, so the last accuracy and memory purity increases and have larger gap as the task number increases.

\begin{figure*}[h!]
    \centering
    \begin{subfigure}[t]{0.25\linewidth}
        \includegraphics[width=\textwidth]{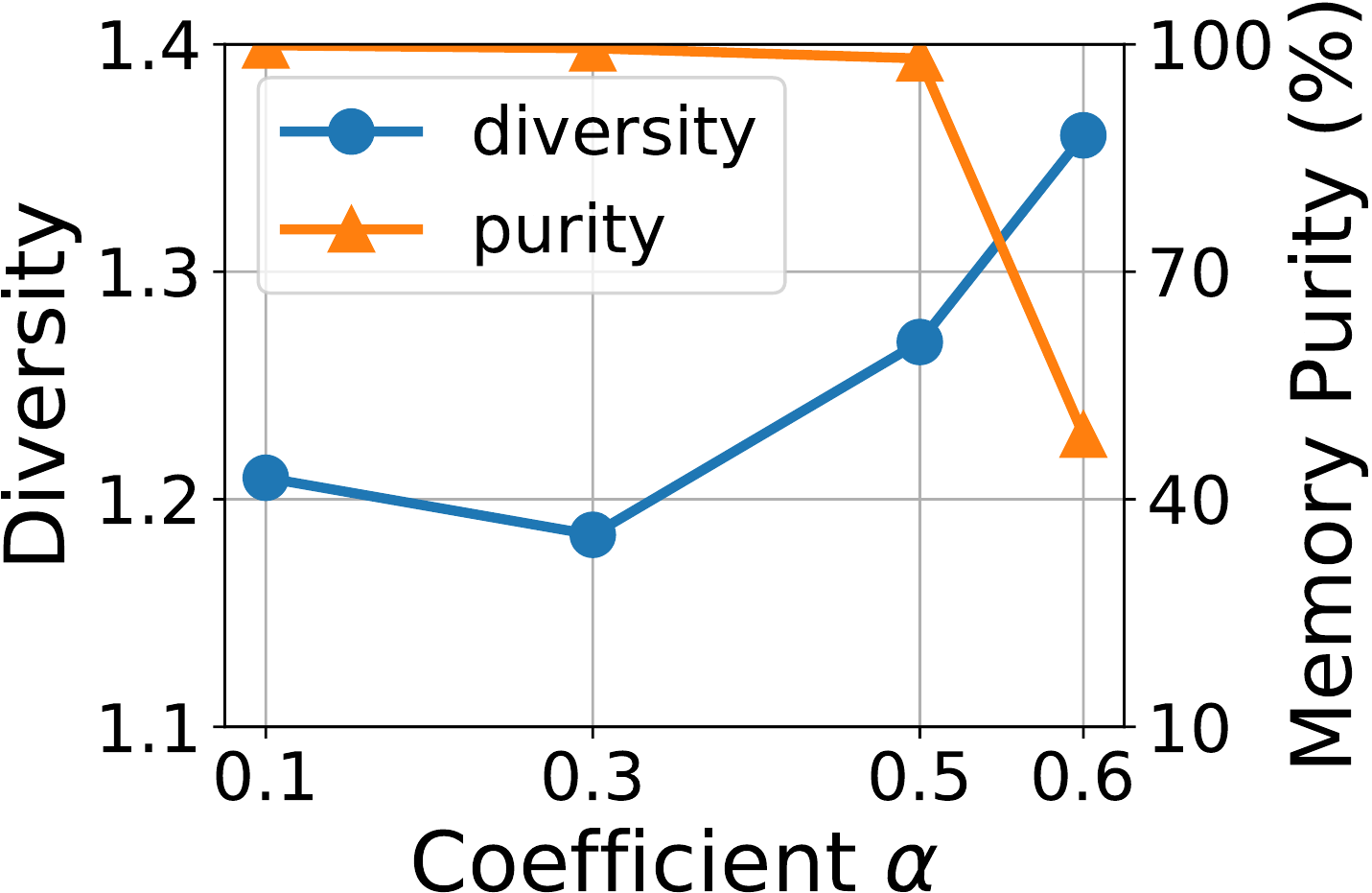}
        \caption{}
        \label{fig:cifar10_sym20_div_pur}
    \end{subfigure}
    \hspace{1em}
    \begin{subfigure}[t]{0.25\linewidth}
        \includegraphics[width=\textwidth]{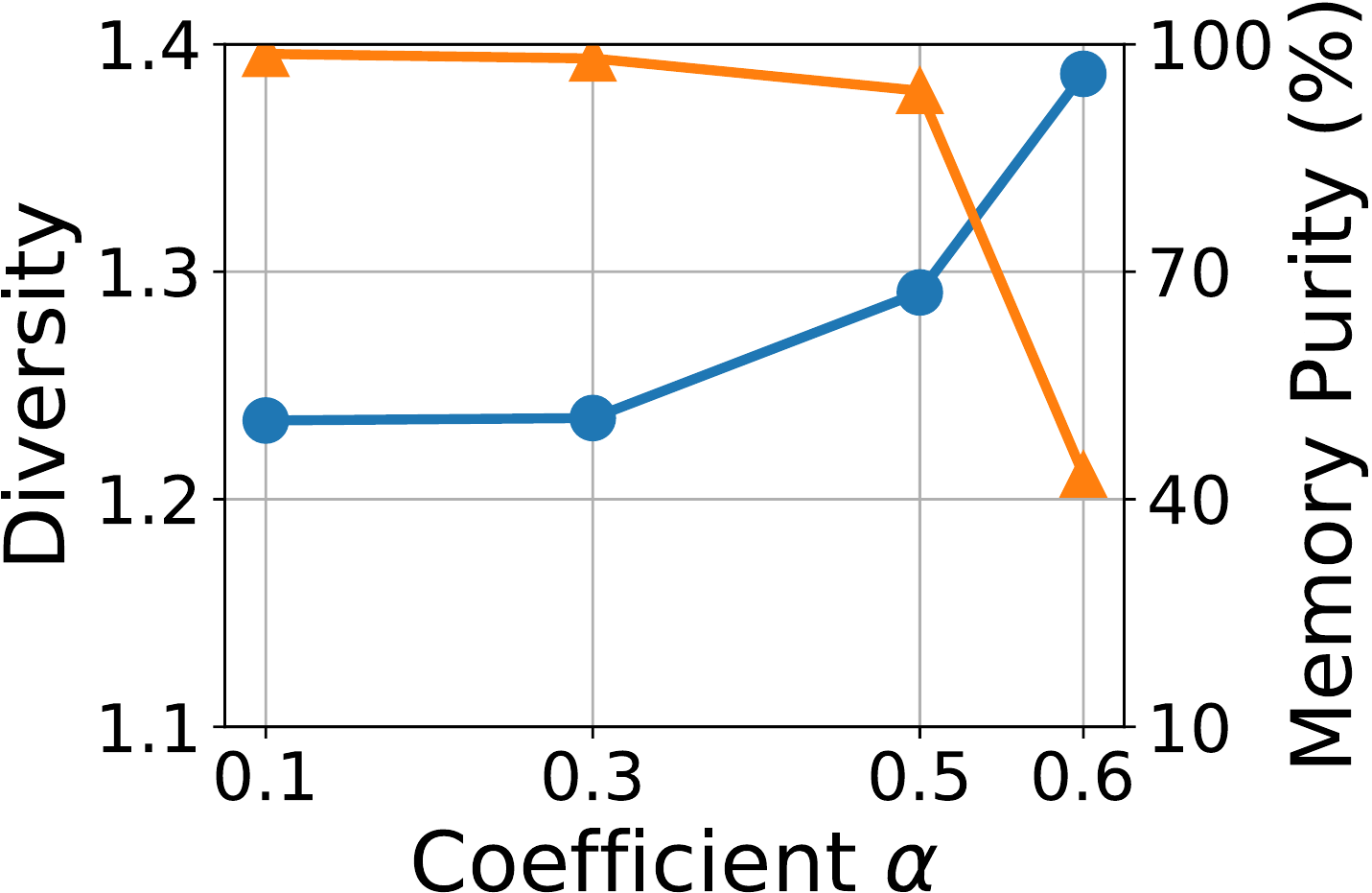}
        \caption{}
        \label{fig:cifar10_sym40_div_pur}
    \end{subfigure}
    \hspace{1em}
    \begin{subfigure}[t]{0.25\linewidth}
        \includegraphics[width=\textwidth]{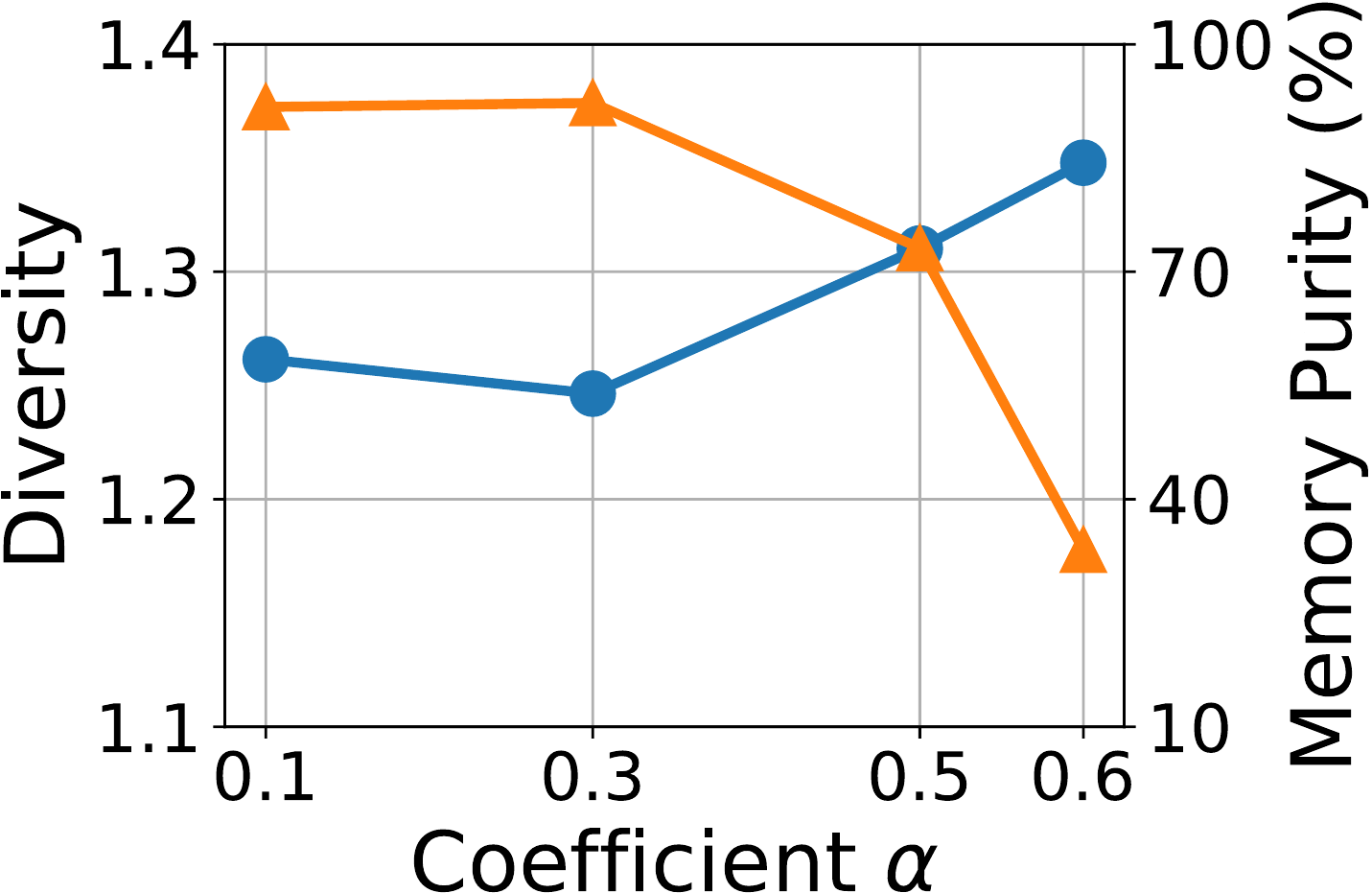}
        \caption{}
        \label{fig:cifar10_sym60_div_pur}
    \end{subfigure}
    
    \caption{Illustration of diversity and purity in the memory as the coefficient $\alpha$ changes on CIFAR-10 with (a) SYM-20\%, (b) SYM-40\%, and (c) SYM-60\%. }
    \label{fig:cifar10_diversity_vs_purity}
\end{figure*}

\begin{figure*}[h!]
    \centering
    \begin{subfigure}[t]{0.25\linewidth}
        \includegraphics[width=\textwidth]{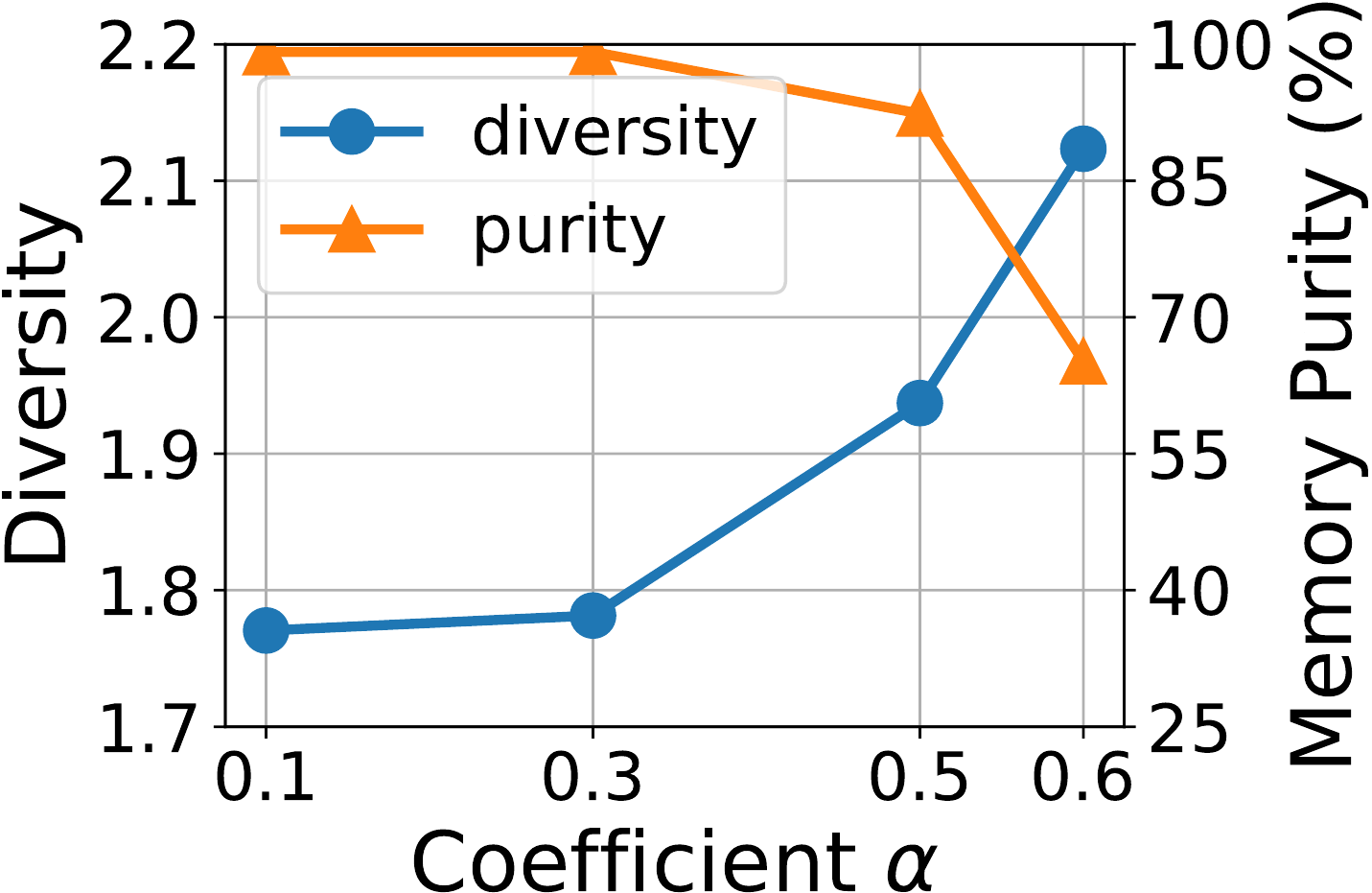}
        \caption{}
        \label{fig:cifar100_sym20_div_pur}
    \end{subfigure}
    \hspace{1em}
    \begin{subfigure}[t]{0.25\linewidth}
        \includegraphics[width=\textwidth]{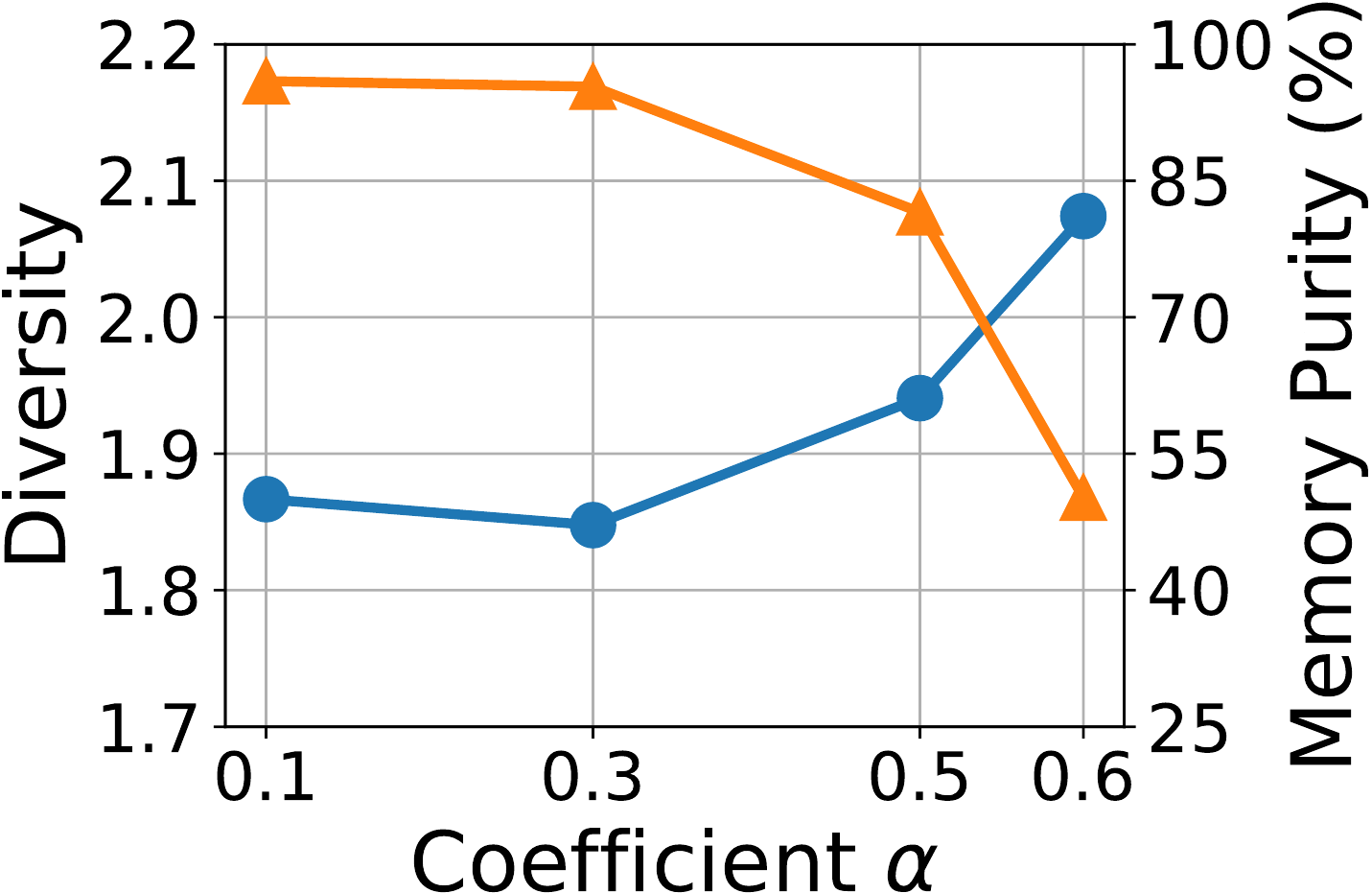}
        \caption{}
        \label{fig:cifar100_sym40_div_pur}
    \end{subfigure}
    \hspace{1em}
    \begin{subfigure}[t]{0.25\linewidth}
        \includegraphics[width=\textwidth]{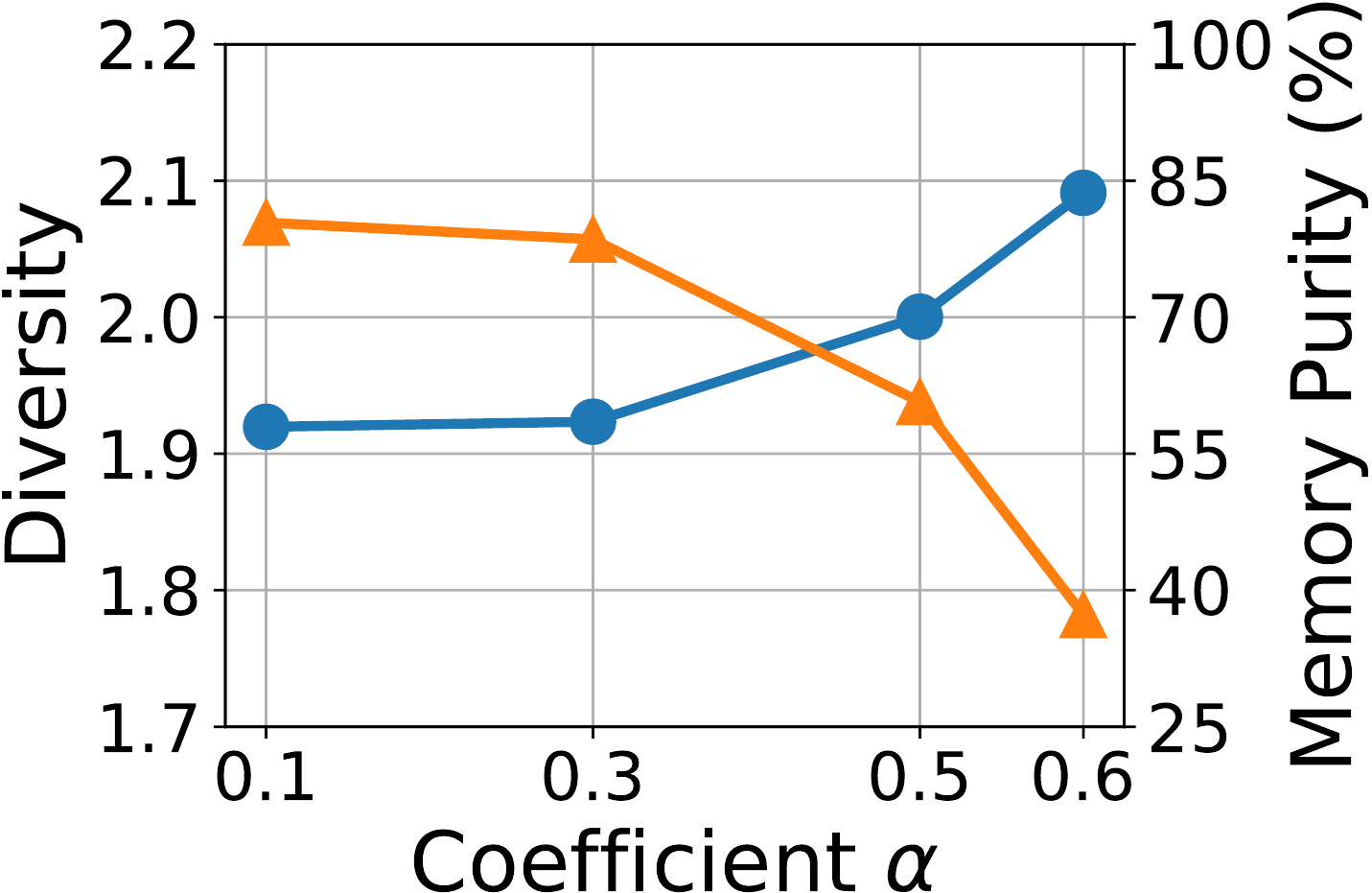}
        \caption{}
        \label{fig:cifar100_sym60_div_pur}
    \end{subfigure}
    
    \caption{Illustration of diversity and purity in the memory as the coefficient $\alpha$ changes on CIFAR-100 with (a) SYM-20\%, (b) SYM-40\%, and (c) SYM-60\%. }
    \label{fig:cifar100_diversity_vs_purity}
\end{figure*}

\section{Diversity vs. Purity}
\label{app:div_pur}
To analyze how diversity and purity in a memory change according to coefficient $\alpha$, we plot the diversity and purity metrics (defined in \bmp{Eq. 12}) in Fig.~\ref{fig:cifar10_diversity_vs_purity} and \ref{fig:cifar100_diversity_vs_purity} for various noise ratios.
As hyper-parameter $\alpha$ increases, the diversity score increases while the purity score decreases in all figures.
When $\alpha$ is increased beyond a certain value, it can be seen that purity is sacrificed for the sample diversity.
If there are too many noisy labels in a memory, a trained model would inevitably learn the noisy labels, which can lead to performance degradation. 
Therefore, the best strategy is to set the $\alpha$ before the abrupt decrease in the purity score.
For example, in Fig.~\ref{fig:cifar10_sym20_div_pur} and \ref{fig:cifar10_sym40_div_pur}, memory purity is over 95\% until $\alpha=0.5$, and then falls to less than 50\% when $\alpha$ is set to 0.6.
Thus, the best strategy for balancing the diversity and memory purity can be obtained at $\alpha=0.5$.
Meanwhile, in the case of Fig.~\ref{fig:cifar10_sym60_div_pur}, memory purity drops to 70\% at $\alpha=0.5$, so it can be expected that the best performance is obtained at $\alpha=0.3$, before memory purity is too much degraded.
This is because memory purity becomes more important as the ratio of noisy labels contained in memory increases.
These results are consistent with the results in \bmp{Tab. 4}.

\begin{table}[t!]

\centering
\caption{Comparison of our method and SPR in disjoint setup.}
\vspace{-1em}
\label{tab:comp_spr}
\resizebox{1.0\linewidth}{!}{%
\begin{tabular}{@{}lrrrrrc@{}}
\toprule 
& \multicolumn{5}{c}{CIFAR-10} & \multicolumn{1}{c}{WebVision} \\
\cmidrule(lr){2-6} \cmidrule(lr){7-7} 
& \multicolumn{3}{c}{Sym.} & \multicolumn{2}{c}{Asym.}  \\ 
Methods  & 
\multicolumn{1}{c}{20} & \multicolumn{1}{c}{40} & \multicolumn{1}{c}{60} & \multicolumn{1}{c}{20} & \multicolumn{1}{c}{40} & Real \\

\cmidrule(lr){1-1} \cmidrule(lr){2-4} \cmidrule(lr){5-6} \cmidrule(lr){7-7} 
Self-Centered filter [\bmp{17}] & 36.5 & 35.7 & 32.5 & 37.1 & 36.9 & 33.0 \\
Self-Replay [\bmp{17}] & 40.1 & 31.4 & 22.4 & 44.1 & 43.2 & 48.0 \\
SPR [\bmp{17}] & 43.9 & 43.0 & 40.0 & 44.5 & 43.9 & 40.0 \\
\cmidrule(lr){1-1} \cmidrule(lr){2-4} \cmidrule(lr){5-6} \cmidrule(lr){7-7} 
\textbf{\our} (Ours) & \textbf{61.2} & \textbf{60.9} & \textbf{56.0} & \textbf{62.4} & \textbf{46.4} & \textbf{51.8} \\
\bottomrule
\end{tabular}
}
\vspace{-0.2em}
\end{table}

\section{Comparison to SPR in Disjoint Tasks}
\label{app:spr}
As SPR [\bmp{17}] is proposed for similar task setup, we compare our \our to SPR [\bmp{17}] in the disjoint setup for which the SPR is proposed, and summarize the results in Tab.~\ref{tab:comp_spr}.
Following their experimental setup, we configure five tasks with randomly paired classes for CIFAR-10, %
and use the top 14 largest classes in seven tasks with randomly paired classes for WebVision. 

SPR uses two different memories; a delayed buffer, which the incoming data is saved temporarily, and a purified buffer which maintains purified data, and a fixed memory size of 500 or 1,000 for CIFAR-10 or WebVision, respectively. 
As the delayed buffer is only utilized to auxiliary model (called Self-Replay Model) to use only purified examples for model training, their actual memory size is $2\times$ of the episodic memory.  
Hence, we set memory size as 1,000 and 2,000 in CIFAR-10 and WebVision for fair comparison.

Interestingly, \our outperforms SPR by 3\%-18\% in all experiments (Tab.~\ref{tab:comp_spr}).
We believe it is because the SPR always purifies incoming contaminated data stream before training the model, but it has a high risk of overfitting by false model predictions.